\pdfoutput=1
\def\isarxiv{1} %
\newif\ifathanorpreprint

\ifdefined\isarxiv
\PassOptionsToPackage{capitalize,noabbrev,sort}{cleveref}
\RequirePackage{colortbl} %
\documentclass{bytedance_seed}
\else

\documentclass{article} %
\usepackage{iclr2027_conference,times}
\ifathanorpreprint
 \iclrfinalcopy %
\fi

\fi

\usepackage{amsmath}
\usepackage{amsthm}

\usepackage{algorithm}
\ifdefined\isarxiv\else\usepackage{subfig}\fi
\usepackage{listings}
\usepackage{upquote}
\usepackage{placeins}
\usepackage{graphicx}
\usepackage{grffile}
\usepackage{wrapfig,epsfig}
\usepackage{fontawesome5} %
\usepackage{url}
\usepackage{xcolor}
\usepackage{colortbl}
\usepackage[most]{tcolorbox}
\definecolor{seedblue}{HTML}{2E5AA8}
\definecolor{seedperiwinkle}{HTML}{5686EE}
\definecolor{seedgreen}{HTML}{4BCF9C}
\definecolor{seedleaf}{HTML}{83D683}
\definecolor{seedrose}{HTML}{E99BAF}
\definecolor{seedlavender}{HTML}{A4B3EE}
\definecolor{seedice}{HTML}{C8EBF6}
\definecolor{seedorange}{HTML}{E05A2B}
\definecolor{seedgray}{HTML}{717071}
\definecolor{agent1}{HTML}{4F72B8}
\definecolor{agent2}{HTML}{3A8F6E}
\definecolor{agent3}{HTML}{B5622E}
\definecolor{agent4}{HTML}{9A4E94}
\newcommand{\agent}[1]{\textcolor{agent#1}{Agent~#1}}
\newcommand{\apt}{\textcolor{seedblue}{AutoPostTrain}}
\colorlet{athanorblue}{seedblue}
\newcommand{\searchvar}[1]{\textcolor{seedperiwinkle!82!black}{#1}}
\newcommand{\modelvar}[1]{\textcolor{seedgreen!52!black}{#1}}
\newcommand{\obsvar}[1]{\textcolor{seedblue}{#1}}
\newcommand{\latentvar}[1]{\textcolor{seedrose!62!black}{#1}}
\newcommand{\feedbackvar}[1]{\textcolor{seedorange!82!black}{#1}}
\newcommand{\policyvar}[1]{\textcolor{seedrose!55!seedblue}{#1}}
\newcommand{\historyvar}[1]{\textcolor{seedgray}{#1}}
\newcommand{\tablehead}{\rowcolor{seedblue!8}}
\newcommand{\keyrow}{\rowcolor{seedgreen!11}}
\newcommand{\fullmark}{\textcolor{seedblue}{\bfseries Y}}
\providecommand{\partmark}{}%
\renewcommand{\partmark}{\textcolor{seedgreen!55!black}{\bfseries P}}
\newcommand{\nomark}{\textcolor{seedgray!70}{--}}
\newtcolorbox[auto counter]{agenttrace}[2][]{
 enhanced,
 breakable,
 colback=seedblue!2,
 colframe=seedblue!35,
 colbacktitle=seedblue!8,
 coltitle=black,
 title={Trace~\thetcbcounter: #2},
 label type=trace,
 fonttitle=\bfseries\small,
 boxrule=0.45pt,
 arc=1.2mm,
 left=1.5mm,
 right=1.5mm,
 top=0.9mm,
 bottom=0.9mm,
 before skip=5pt,
 after skip=5pt,
 #1
}
\lstdefinestyle{agentnote}{
  upquote=true,
 language={},
 basicstyle=\scriptsize\ttfamily,
 breaklines=true,
 breakatwhitespace=true,
 columns=fullflexible,
 keepspaces=true,
 backgroundcolor=\color{seedblue!3},
 frame=single,
 rulecolor=\color{seedblue!18},
 framerule=0.4pt,
 framesep=3pt,
 xleftmargin=4pt,
 xrightmargin=4pt,
 aboveskip=4pt,
 belowskip=6pt,
}
\lstdefinestyle{athanorpython}{
 language=Python,
 basicstyle=\scriptsize\ttfamily,
 keywordstyle=\color{seedblue}\bfseries,
 commentstyle=\color{seedgray}\itshape,
 stringstyle=\color{seedgreen!55!black},
 emph={research,inspect,submit,note,open_ledger,decision_due,research_view,start_async,received_requests,execution_updates,handle_request,ready_work,dispatch},
 emphstyle=\color{seedperiwinkle!70!black},
 backgroundcolor=\color{seedblue!3},
 frame=single,
 rulecolor=\color{seedblue!18},
 framerule=0.4pt,
 framesep=3pt,
 columns=fullflexible,
 keepspaces=true,
 showstringspaces=false,
 breaklines=true,
 aboveskip=0pt,
 belowskip=0pt
}
\usepackage{epstopdf}
\usepackage{enumitem}

\usepackage{bbm}
\usepackage{dsfont}

\usepackage{booktabs}
\usepackage{tabularx}
\usepackage{longtable}
\usepackage{needspace}

\allowdisplaybreaks

\ifdefined\isarxiv

\usepackage{amssymb}
\makeatletter
\DeclareTextFontCommand{\textbf}{\fontsize{\strip@pt\dimexpr0.88\dimexpr\f@size pt\relax\relax}{\f@baselineskip}\selectfont\bfseries\sffamily}
\makeatother

\else

\usepackage{amssymb} %

\usepackage{microtype}
\usepackage{hyperref}
\definecolor{mydarkblue}{rgb}{0,0.08,0.45}
\hypersetup{colorlinks=true, citecolor=mydarkblue,linkcolor=mydarkblue, pdftitle={Athanor: An Agent-Native System for Autonomous Post-Training Research}, pdfauthor={Yufa Zhou, Yichen Wang, Jiakun Fan, Ziqi Shu, Zhichao Liu, Anru R. Zhang, Neiwen Ling, Zijun Wei}}

\usepackage[capitalize,noabbrev,sort]{cleveref} %

\fi

\theoremstyle{plain}

\makeatletter
\newcommand*{\RN}[1]{\expandafter\@slowromancap\romannumeral #1@}
\makeatother

\AtBeginDocument{\hypersetup{pdftitle={Athanor: An Agent-Native System for Autonomous Post-Training Research}}}
\ifdefined\isarxiv\AtBeginDocument{\hypersetup{pdfauthor={Yufa Zhou, Yichen Wang, Jiakun Fan, Ziqi Shu, Zhichao Liu, Anru R. Zhang, Neiwen Ling, Zijun Wei}}}\fi
\crefname{trace}{Trace}{Traces}\Crefname{trace}{Trace}{Traces} %
\begin{document}

\title{
Athanor: An Agent-Native System\\ for Autonomous Post-Training Research
}

\ifdefined\isarxiv
\author[1,2,\ddagger,\dagger]{Yufa Zhou}
\author[3]{Yichen Wang}
\author[1,4]{Jiakun Fan}
\author[1]{Ziqi Shu}
\author[1]{Zhichao Liu}
\author[2]{\\Anru R. Zhang} %
\author[1,\dagger]{Neiwen Ling}
\author[1,\dagger]{Zijun Wei}
\affiliation[1]{ByteDance}
\affiliation[2]{Duke University}
\affiliation[3]{The University of Chicago}
\affiliation[4]{Virginia Tech}
\contribution[\ddagger]{Project lead}
\contribution[\dagger]{Corresponding authors}
\correspondence{Yufa Zhou at \email{yufa.zhou@duke.edu}, Neiwen Ling at \email{neiwen.ling@bytedance.com}, Zijun Wei at \email{zijun.wei@bytedance.com}}
\abstract{Recursive self-improvement, in which AI systems build stronger AI systems, is
one promising route to superintelligence. We study a weaker form that can be
measured today: frontier agents post-training a small raw base model with
public data, as a test of their research potential. We build
\textbf{Athanor}, an open agent-native system for \textbf{autonomous
post-training research} (\apt) on real GPUs. We give the agents \textit{full research
autonomy}, low-friction engineering tools, and efficient starter code. The agents carry out the research themselves: they read
datasets and evaluation results, build their training data, write and modify
training algorithms, tune hyperparameters, and analyze the results to decide
what to try next. They work in a sandbox and cannot use external teacher models. Athanor's
services supply frozen evaluation, managed jobs, decontaminated data, a
persistent research record, and a held-out suite the agents are unaware of. In
a 104-hour campaign, a team of four Claude Opus 5.5 agents, each on an
eight-GPU node and given no research advice, improves Qwen3-4B-Base from
\textbf{15.9} to \textbf{67.3} on a six-task development suite and from \textbf{9.2} to
\textbf{48.2} on a held-out suite, against \textbf{66.5} and \textbf{47.9} for the official
Qwen3-4B-Instruct-2507. To our knowledge, this is the
first demonstration that a team of frontier agents can autonomously
post-train a raw base model into a cross-domain generalist model that \textbf{matches or
surpasses the official instruct version} on average.
The campaign's full record shows that the agents reach this level by recombining known methods and
each other's models: SFT on public reasoning data, RL with a rollout change one
agent introduced, weight merges across agents, and rounds of
self-distillation. With execution handled, today's agents already have the engineering and research ability to conquer this task.
We hope Athanor provides a foundation for scaling autonomous AI research.

\par\vspace{1.2em}

\begin{flushright}
\begin{minipage}{0.8\textwidth}
\small\itshape
``Instead of trying to produce a programme to simulate the adult mind, why not
rather try to produce one which simulates the child's? If this were then
subjected to an appropriate course of education one would obtain the adult
brain.''\par\vspace{2pt}
\raggedleft\upshape --- Alan M.\ Turing, \textit{Computing Machinery and Intelligence}, 1950
\end{minipage}
\end{flushright}

\vspace{2pt}
}
\else
\date{}

\author{%
{\bfseries Yufa Zhou$^{1,2,\ddagger,\dagger}$ \quad Yichen Wang$^{3}$ \quad Jiakun Fan$^{1,4}$ \quad Ziqi Shu$^{1}$ \quad Zhichao Liu$^{1}$} \\[0.15em]
{\bfseries Anru R. Zhang$^{2}$ \quad Neiwen Ling$^{1,\dagger}$ \quad Zijun Wei$^{1,\dagger}$} \\[0.35em]
{\normalfont\normalsize $^{1}$ByteDance \quad $^{2}$Duke University \quad $^{3}$The University of Chicago \quad $^{4}$Virginia Tech} \\[0.15em]
{\normalfont\small\texttt{yufa.zhou@duke.edu}\quad\texttt{\{neiwen.ling, zijun.wei\}@bytedance.com}}
}
\fi

\newcommand{\correspondingauthornote}{%
 \begingroup
 \renewcommand{\thefootnote}{\fnsymbol{footnote}}%
 \footnotetext[3]{Project lead.}%
 \footnotetext[2]{Corresponding authors.}%
 \endgroup
}

\ifdefined\isarxiv
 \maketitle %
 \suppressfloats[t] %

\else

\maketitle
\ifathanorpreprint
 \lhead{Preprint}
 \correspondingauthornote
 \vspace{-\baselineskip}
\fi

\begin{abstract}
 
\end{abstract}

\fi

\begingroup
\setlength{\intextsep}{4pt}
\def\headlinewidth{\textwidth}
\ifdefined\isarxiv\else
 \ifathanorpreprint
 \captionsetup{font=small,skip=3pt}
 \def\headlinewidth{0.94\textwidth}
 \fi
\fi
\ifdefined\isarxiv
 \begin{figure}[!t]
\else
 \begin{figure}[H]
\fi
 \centering
 \includegraphics[width=\headlinewidth]{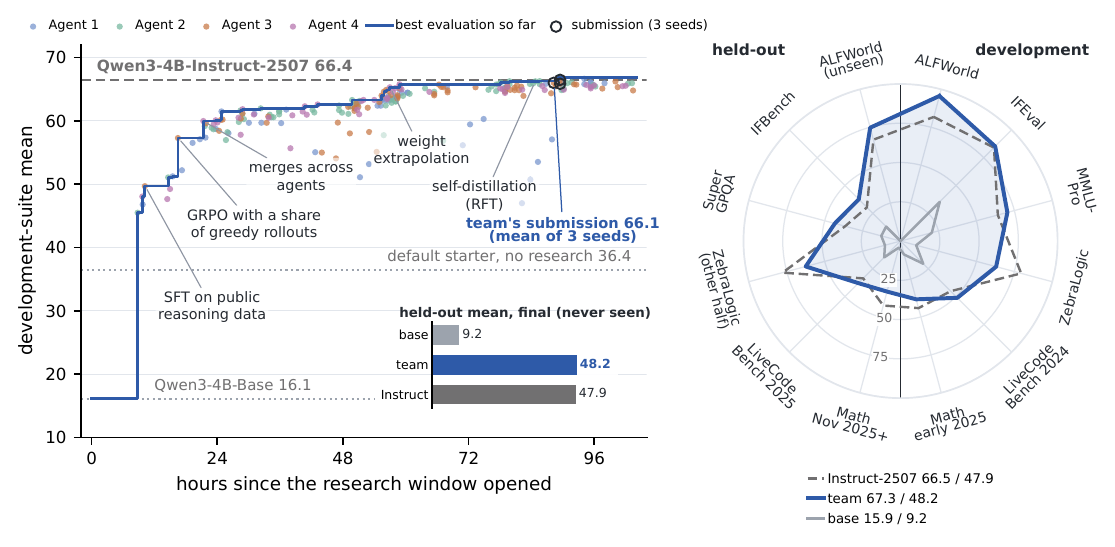}
 \caption{\textbf{A team of four agents post-trains Qwen3-4B-Base to the level of the vendor's instruct release in 104 hours.} The team's submission scores 67.3 against 66.5 for Qwen3-4B-Instruct-2507 on the development suite, and 48.2 against 47.9 on a held-out suite that the agents never see. Left: the search on the development suite under the campaign protocol, the one the agents worked with (\cref{sec:setting}). Each dot is one full evaluation of a checkpoint; the staircase is the best so far. Under this protocol the team's submission averages 66.1 over three evaluation seeds, against 66.4 for Instruct-2507. RFT is rejection-sampling fine-tuning; the default starter is the provided starter code run unchanged. Inset and right: the final measurement, a remeasurement after the campaign with a longer output limit and one sampling setting for every model (\cref{tab:c1-main}); the radar mirrors each development task's held-out twin on the left.}
 \label{fig:c1-climb}
\end{figure}
\endgroup

\raggedbottom
\ifdefined\isarxiv\newpage\fi %
\def\SystemGlanceFigure#1{%
\begingroup
\setlength{\intextsep}{6pt}
  \begin{figure}[#1]
  \centering
  \includegraphics[width=\textwidth]{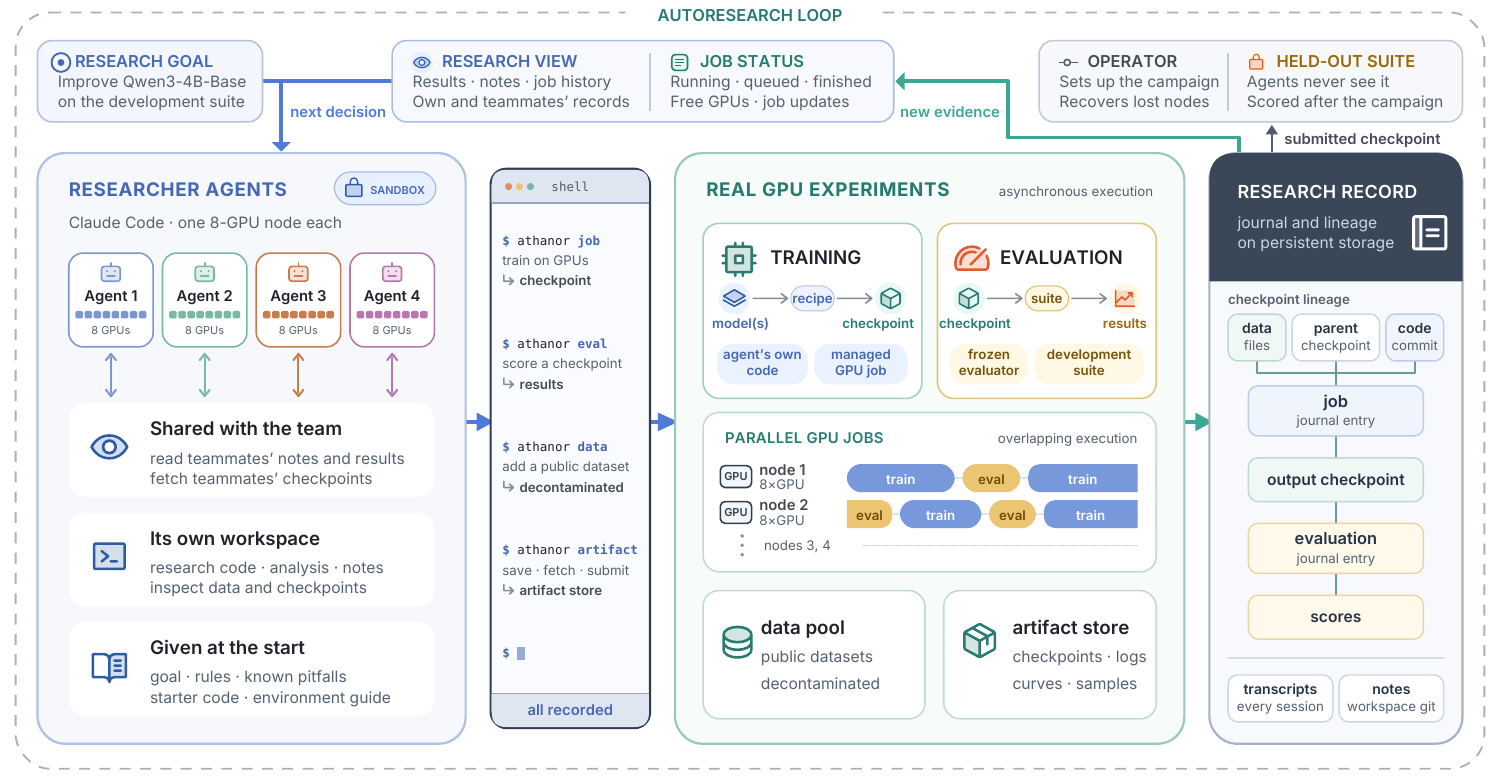}
  \caption{\textbf{Athanor at a glance.} Each researcher agent works in its own sandbox on a GPU node, with its own coding tools, and reaches the GPUs through one command: \texttt{athanor job}, \texttt{eval}, \texttt{data}, and \texttt{artifact} run jobs, request evaluations, bring in data, and save or submit checkpoints. The environment runs the jobs, decontaminates incoming data, scores checkpoints with a frozen evaluator, and records every request and result for the team. The operator, who runs the campaign, sets it up, recovers lost nodes, and scores the team's submission on the held-out suite.}
  \label{fig:system-at-a-glance}
\end{figure}
\endgroup}
\ifdefined\isarxiv\else\SystemGlanceFigure{H}\fi

\section{Introduction}\label{sec:intro}

Recursive self-improvement (RSI), in which an AI system uses its own capabilities to build a more capable successor~\citep{good1966,schmidhuber2007godel}, is one of the proposed routes to superhuman intelligence~\citep{genewein2026agi}. Its early forms are already reported~\citep{anthropic2026rsi,amodei2026pace}: AI systems write most of the code inside the companies that build them~\citep{chan2026explosion}, and agents carry out a growing share of the research~\citep{openai2026researchacceleration}.
  For language models, the part of this loop open to study is post-training: pretraining establishes the knowledge foundation~\citep{kaplan2020scaling,hoffmann2022training,llama3},
while post-training shapes instruction following, reasoning, coding, and
alignment~\citep{rlhfbook,openai_o1,tulu3,qwen3,guo2025deepseekr1}. Yet
post-training research remains an expensive iterative search, and published
reports typically document the final recipe rather than the full research trajectory that
produced it~\citep{qwen3.5,nemotron,xu2026deepseekv4,gong2026researchtrails}. Search over data mixtures, objectives, algorithms, and hyperparameters is an
evidence-driven loop that agents have begun to
automate~\citep{aiscientist,alphaevolve,chen2026automatedalignment} and
benchmarks have begun to evaluate~\citep{mlebench,rebench,mlsbench}.

\ifdefined\isarxiv\SystemGlanceFigure{!t}\fi
We study \textbf{autonomous post-training research} (\textbf{\apt}), where a frontier agent
diagnoses a base model's weaknesses, runs training experiments at scale, and
iteratively refines its recipes from the evidence. Existing work studies narrower versions of this problem. PostTrainBench and Locus have agents train a separate \textit{specialist} for each benchmark~\citep{ptb,locus}, and other benchmarks and systems isolate fine-tuning, reinforcement-learning engineering, data-centric research, or structured agent workflows~\citep{ftdojo,agent2rlbench,rsibenchdata,du2026datamaster,yu2026autotrainess}. These works also allow teacher models or do not keep a held-out evaluation sealed from the agents.
 Our setting takes the harder choice on every one of these points: one \textit{generalist} checkpoint must improve on six capabilities at once, starting from a raw base model, with no teacher model, and it is scored on a held-out suite that the agents never see, which tests the same capabilities on other questions.

\apt{} is expensive: one hypothesis can take hours of failure-prone GPU execution before its checkpoint can be evaluated. It is also hard to keep honest: a search that runs for days must keep its training data clean and its final evaluation sealed~\citep{contamination,rewardhacking}.
 For an agent to do this
research, experiments must be easy to submit, the evaluation must stay out of the agent's reach, and, because the search outlasts any one session, the research state must be
\textit{externalized}: kept outside the agent's context, with each result
linked to the evidence then available and the recipe that produced it.

We build \textbf{Athanor} (\cref{fig:system-at-a-glance}), an open
agent-native system that enables frontier agents to hill-climb a raw base model
through long-running campaigns of training experiments on real GPUs. Our design
thesis is that, \textit{as agent intelligence improves, the bottleneck in
autonomous research shifts from generating plausible ideas to the friction of
executing and verifying them}: without managed execution, an agent spends hours debugging training infrastructure and rebuilds the same machinery for every experiment. Athanor therefore makes post-training research \textit{programmable}: it gives research its atomic operations (starting a job, evaluating a checkpoint, bringing in data), the agent composes them in its own code, and the environment carries them out. The system is agent-native in that the agent does research with its own coding and analysis tools. The central design question is \textit{how to
empower agents, not imprison them}: the agents keep every research decision, while experiment execution stays \textit{efficient, auditable, and
recoverable}.

Three design principles make this work. \textbf{First, separate research decisions from execution:} the agent inspects data and model behavior, writes its own data and training programs, and decides what to run; the environment runs, measures, and recovers every experiment. \textbf{Second, externalize the research state:} every request, result, checkpoint lineage, transcript, and note stays in a durable record, so a session can be resumed or replaced, several agents can work as a team~\citep{park2026communication}, and every campaign can be audited. \textbf{Third, keep the research loop productive:} jobs run in the background and each result reaches the agent the moment it arrives, so an agent keeps several experiments in flight.

We evaluate Athanor in a \textbf{large-scale post-training} campaign. Four
agents (Claude Opus 5.5 in Claude Code~\citep{opus55}) form a team, each on one
node of eight GPUs, and start from Qwen3-4B-Base with public data, no
teacher-model calls, and no research advice from humans. In 104 hours the team raises
the base from 15.9 to 67.3 on a six-task development suite and from 9.2 to 48.2
on a held-out suite of different questions for the same capabilities, which the
agents never know exists; the vendor's Qwen3-4B-Instruct-2507\footnote{\url{https://huggingface.co/Qwen/Qwen3-4B-Instruct-2507}, the vendor's July 2025 instruction-tuned update of Qwen3-4B.} scores 66.5 and 47.9 under the same evaluator (\cref{fig:c1-climb}, \cref{tab:c1-main}).
To the best of our knowledge, this is \textit{the
first public result in which a team of AI agents, working autonomously, post-trains a raw base model into one
cross-domain generalist model at the level of the vendor's instruct release, confirmed on the held-out suite.} Earlier results at this level come from a human team's recipe, T\"ulu~3, which generated training
data with stronger models~\citep{tulu3}, or from agents that train one specialist per benchmark or call teacher
models~\citep{locus,du2026rsimaster}; other agent systems report gains on tasks seen during development or on one task family~\citep{meng2026rsigym,shi2026evolve}. \Cref{sec:related_work} and \cref{app:related-tables} compare the agent systems.
The vendor describes building its small Qwen3 models by distillation from its own larger flagship models~\citep{qwen3}; the agents reach the same level from public data, much of which was itself distilled from stronger models by its publishers~\citep{amthinking2025,tulu3,nemotron}.

We then analyze the full record of the campaign, including the agents' transcripts and notes, and find (\cref{sec:discoveries}):
(1) \textit{recombination}: the recipe chains known methods (supervised fine-tuning, reinforcement learning, weight merging, and self-distillation), and the agents contribute the composition and the small changes that make it work;
(2) \textit{team progress}: ideas and checkpoints move between agents within hours, and every new best after hour 21 combines the training of several agents; (3) \textit{recipe convergence}: within the first hour of the campaign, all four agents build on the same public corpus of reasoning traces, and the alternatives they try later score lower and are dropped; and (4) \textit{uneven data sense}: the agents catch problems of format, length, scorability, and difficulty in their data but rarely check whether its answers are correct.
Taken together, the agents' engineering, planning, reasoning, and knowledge of post-training already suffice to take a raw base model to the level of the vendor's instruct release. Once the friction of execution is removed, the clearest limit we observe is how many different ideas they try: four copies of one agent model start from the same prior, converge on one recipe early, and test few alternatives. How the output of this research, the best checkpoint and the understanding of post-training behind it, grows with the compute the agents spend is what we call \textbf{autoresearch scaling} (\cref{sec:autoresearch}).

In summary, this paper makes three contributions:
\begin{itemize}[leftmargin=1.4em,itemsep=1pt,topsep=3pt]
  \item \textbf{Athanor}, an open agent-native system for post-training
  research on real GPUs, with frozen evaluation, managed jobs, decontaminated
  data, and a research record that links every score to the recipe behind it.
  \item \textbf{Instruct-level post-training by a team of agents}: in 104 hours, and for about \$12,000 in all (GPU time at
  rental prices and the agents' inference), four agents post-train Qwen3-4B-Base into one
  cross-domain generalist model that matches or surpasses the vendor's instruct release (67.3 against
  66.5 on the development suite and 48.2 against 47.9 on a held-out suite),
  without a teacher model.
  \item \textbf{Four findings on how the agents do research}, from the campaign's full record (500 jobs, 255 requested full evaluations, and the lineage of every checkpoint): they recombine known methods, progress as a team, converge early on one recipe, and inspect the form of their data but rarely its correctness.
\end{itemize}

\section{On AutoPostTrain}\label{sec:autoresearch}

\subsection{Problem Formulation}

Given a base model and an experimental budget, \apt{} searches over executable recipes $\searchvar r$, each specifying
data, objective, method and program, hyperparameters, stage order, and
initial checkpoint. Executing the recipe yields a checkpoint
$\modelvar M=\operatorname{Exec}(\searchvar r)$. Since a recipe may start from
any completed checkpoint, the search branches into lineages that compose
supervised fine-tuning (SFT), reinforcement learning with verifiable rewards, on-policy distillation, and merging~\citep{rlhfbook,tulu3,qwen3,guo2025deepseekr1,nemotron}. Let
$\modelvar{\mathcal P}_{t-1}$ denote the pool of checkpoints available before
decision $t$, with $\modelvar{\mathcal P}_0$ containing only the base model,
and let $\searchvar{\mathcal R}(\modelvar{\mathcal P}_{t-1})$ denote the
recipes admissible on this pool. Let the search objective $\latentvar c(\modelvar M)$ denote the
latent capability of checkpoint $\modelvar M$; it
is not directly observable and is estimated by evaluation on a finite
benchmark suite $\obsvar{\mathcal B}$. A researcher agent acts as a policy
$\policyvar\pi$ over recipes, conditioned on the history
$\historyvar{\mathcal H}_{t-1}$ of recipes, programs, measurements, failures,
and notes. The \apt{} loop is:
\begin{equation}
\tag{\apt}\label{eq:recipe-search}
\begin{aligned}
\text{\color{seedgray}\scriptsize RESEARCH}\quad
&\searchvar{r_t}\sim\policyvar\pi(\cdot\mid\historyvar{\mathcal H}_{t-1}),
\qquad \searchvar{r_t}\in\searchvar{\mathcal R}(\modelvar{\mathcal P}_{t-1}),
\\[1pt]
\text{\color{seedgray}\scriptsize EXPERIMENT}\quad
&\modelvar M_t=\operatorname{Exec}(\searchvar{r_t}),\qquad
  \obsvar{\widehat{\mathbf p}_t}
  =\operatorname{Eval}_{\obsvar{\mathcal B}}
    (\modelvar M_t;\feedbackvar{\xi_t}),
\\[1pt]
\text{\color{seedgray}\scriptsize OBJECTIVE}\quad
&\modelvar M_T^\star\in
  \underset{\modelvar M\in\modelvar{\mathcal P_T}}
  {\operatorname*{arg\,max}}\ \latentvar c(\modelvar M).
\end{aligned}
\end{equation}

At step $t$, the policy proposes a recipe, execution produces a checkpoint,
and evaluation returns its benchmark score vector
$\obsvar{\widehat{\mathbf p}_t}$, where $\feedbackvar{\xi_t}$ denotes
evaluation randomness. The pool grows as
$\modelvar{\mathcal P_t}=\modelvar{\mathcal P}_{t-1}\cup\{\modelvar M_t\}$,
and the result and notes enter $\historyvar{\mathcal H_t}$. After $T$
completed experiments, the objective $\modelvar M_T^\star$ is the checkpoint of highest latent capability in the final pool, whereas the researcher observes only $\obsvar{\widehat{\mathbf p}_t}$. Because
$\obsvar{\widehat{\mathbf p}_t}$ holds one score per benchmark, a recipe can
raise one capability while lowering another, which makes the observed search
\textit{multi-objective}~\citep{sener2018multi}. Several researchers can run this loop at once as a \textit{team}: researchers $\policyvar{\pi_1},\dots,\policyvar{\pi_K}$ each propose recipes, $\searchvar{r_t^{(k)}}\sim\policyvar{\pi_k}(\cdot\mid\historyvar{\mathcal H}_{t-1})$, from one shared history $\historyvar{\mathcal H}$ and one shared pool $\modelvar{\mathcal P}$, so each reads the others' recipes, results, and notes, and may start from their checkpoints.

\subsection{What Makes AutoPostTrain Hard}

\paragraph{Combinatorial, state-dependent search.}
In the RESEARCH step of \eqref{eq:recipe-search}, the agent chooses a recipe
from a high-dimensional, compositional space: discrete choices of data and
method, continuous hyperparameters, and conditional training stages. Gradient descent optimizes parameters within a stage;
the researcher decides the order of training stages and their starting
checkpoints. The choice is state-dependent: each
stage trains the checkpoint produced by earlier stages, so upstream stages
condition what later stages can achieve~\citep{zhang2025interplay},
and each new checkpoint enlarges the admissible set
$\searchvar{\mathcal R}(\modelvar{\mathcal P_t})$. Since the benchmark scores that serve as feedback are not differentiable with respect to the recipe, the
researcher can only propose a promising
recipe, observe a noisy score, and keep the change if the score improves: the outer
loop is \textbf{stochastic black-box hill climbing}~\citep{russell2021aima}.

What lets an agent do better than a blind hill climber is its prior and the feedback it can reason over. A researcher agent's pretrained knowledge acts as a strong \textit{scientific prior} that
concentrates proposals in a small effective region of the recipe space, and it can reason about why a change helped. To
choose the next recipe within that region, it needs feedback it can act on. Because a recipe's effect depends on its starting
checkpoint, each score is useful only if it is linked to the full lineage that produced it.

\paragraph{Delayed, noisy, and incomplete feedback.}
The EXPERIMENT step of \eqref{eq:recipe-search} returns a checkpoint and its
full benchmark evaluation results. Post-training has no cheap proxy for this result: NanoGPT speedrunning can compare runs by a common validation
cross-entropy target~\citep{modded_nanogpt_2024}, but post-training cannot,
because its stages optimize losses on different scales, and even in pretraining, similar loss values
can correspond to different downstream performance~\citep{liu2023same,chen2026nexus}. To evaluate a
recipe, we must first train it, so there is a delay
$\feedbackvar{\tau_{\mathrm{fb},t}}=
\feedbackvar{\tau_{\mathrm{exec}}}(\searchvar r_t)+
\feedbackvar{\tau_{\mathrm{eval}}}(\modelvar M_t,\obsvar{\mathcal B})$, the time
to prepare data and train \textit{plus} the time to evaluate the checkpoint on
$\obsvar{\mathcal B}$.
The score is noisy: finite evaluation can blur the order of nearby
candidates~\citep{madaan2024quantifying}. It is also incomplete: each benchmark covers only part of its capability, and
IFEval's verifiable constraints, for example, cover only part of instruction following~\citep{zhou2023instruction}.

\paragraph{Scaling the research process.}
Pretraining scaling laws relate loss to compute for a fixed training
procedure~\citep{kaplan2020scaling,hoffmann2022training}, and similar curves
can be fit for RL post-training once the recipe is fixed~\citep{khatri2026art}. If the recipe changes, no stable law is known that relates post-training compute to downstream performance. Here the recipe is what the agent searches for, so we ask how the output of the research grows as the agent
completes more experiments and uses more compute: the best
checkpoint's capability $\latentvar c(\modelvar M_T^\star)$, and the understanding of post-training that the experiments leave behind. We call this relationship
\textbf{autoresearch scaling}. The process spends two resources: agent inference \citep{openai2026researchacceleration} pays for the analysis that invents new recipes, and training and evaluation compute turns those recipes into research artifacts, namely a growing set of checkpoints
and the scientific trajectory that produced them; \cref{fig:c1-climb,fig:c1-cost} show the capability curve against hours and cost, and \cref{sec:discoveries} discusses why it flattens.

\Needspace{17\baselineskip}
\section{Athanor}\label{sec:athanor}

\subsection{The Autoresearch Loop}\label{sec:athanor-loop}

\begin{wrapfigure}[14]{r}{0.45\columnwidth}
  \vspace{-1.0em}
  \captionsetup{type=algorithm,font=small,skip=3pt}
  \caption{The Athanor autoresearch loop. The two loops run concurrently over one shared history.}
  \label{alg:autoresearch}
  \lstset{style=athanorpython}
\begin{lstlisting}
history = persistent_record()

# Each researcher of the team
while research_open():
    evidence = inspect(history)  # own, teammates'
    action = agent.develop(evidence)
    athanor(action)  # job | eval | data | artifact

# Environment services (run independently)
while execution_open():
    updates = collect_job_updates()
    history.record(updates)
    # Train, evaluate, decontaminate, store.
    for job in ready_work(history):
        launch_async(job)
\end{lstlisting}
  \vspace{-0.3em}
\end{wrapfigure}

Athanor implements the loop \eqref{eq:recipe-search} as two
loops over one shared history (Algorithm~\ref{alg:autoresearch} and
\cref{fig:system-at-a-glance}): a researcher loop run by each agent of the
team, and the environment's services. The shared \texttt{history} is the research
record (\cref{fig:system-at-a-glance}, right) of requests, checkpoints,
measurements, and notes. In the researcher loop, each researcher agent (left) reads the record, both its own entries and its teammates' (\texttt{inspect} in Algorithm~\ref{alg:autoresearch}). It analyzes what it
reads in its workspace with its own shell and code tools, and develops the next
step. It then acts through the main verbs of the \texttt{athanor} command
(\texttt{job}, \texttt{eval}, \texttt{data}, and \texttt{artifact};
\cref{fig:system-at-a-glance}).
The command returns at once, so the researcher does not wait for the job it
started. The environment's services (center) run each requested job on the
GPUs when its inputs and resources are ready, write every request and every
result into the same history, and send each result to the researcher as soon
as it arrives.

\subsection{The Research Environment and Its Roles}\label{sec:athanor-philosophy}

\paragraph{The post-training research environment.}
We study a generalist setting in which one checkpoint must improve across a
suite of target capabilities, starting from a raw base model. Our environment
gives each researcher a coding workspace on a GPU node; a pool of public training corpora decontaminated against the benchmarks; and the base model and the
checkpoints saved by its team. Starter code, which the researcher may edit or replace, covers building training datasets, supervised
fine-tuning, group relative policy optimization (GRPO), reinforcement learning in interactive environments, on-policy distillation from a team checkpoint, generating logic puzzles, replaying public ALFWorld expert trajectories, and checkpoint merging, on the verl
trainer~\citep{sheng2025hybridflow} and mergekit~\citep{mergekit}. To test research ability under controlled
conditions, the researcher cannot call or download an external teacher model,
and any further public dataset it brings in goes through the same
decontamination first. The evaluation is protected from the researchers. On
request, the development suite returns every benchmark score and every scored
answer as feedback. A held-out suite, unknown to the researchers, gives the result after the campaign.

\paragraph{No human-designed research scaffolding.}
Athanor does not prescribe how the research is done. Following the bitter
lesson~\citep{sutton2019bitter}, it builds no research procedure into the
agent: the environment does not choose a direction, approve an experiment, or interpret a result, and the researchers run the harnesses that the model developers ship (Claude Code or Codex; this campaign uses Claude Code) through a shell, files, and one command-line tool. In our experience, earlier versions of Athanor that prescribed a form per recipe, an approval per experiment, and a central scheduler constrained stronger agents more than they helped them. Others building
agent systems report related effects: scaffolding written for a weaker model
can constrain the judgment of a stronger one~\citep{lmsys2026inferforge}, and in one study prescribed roles made little difference to what agents produced~\citep{anthropic2026multiagent}. What we improve between campaigns is therefore the environment, not a research procedure: after each campaign we read the agents' trajectories for points of friction, such as slow steps, confusing commands, and recurring workarounds, and improve the tools without changing the researcher's role. Building Athanor is thus a meta loop around the research loop.

\paragraph{Researchers, environment, and operator.}
Athanor separates three roles by what each may decide and what each may see. The
\emph{researchers} make every research decision; they see their own and their teammates' records and the development results, but not the held-out suite, and they cannot change the evaluator. Because the researchers are the ones being studied, these limits are enforced by the software: the sandbox, the frozen evaluator, and the sealed held-out suite. The \emph{environment} makes no research decision: it runs what is requested, measures with the frozen evaluator, and records what happened. The \emph{operator}, a person or an agent acting for one, has the widest authority and sees everything: it sets the campaign's goal and measurement protocol before the start, decides how long the campaign runs, keeps the nodes running, and scores the held-out suite. It gives no research advice, and every operator command leaves a record.

\subsection{Keeping the Loop Fast and Reliable}\label{sec:athanor-provides}

\paragraph{Separate research decisions from execution.}
In the terms of \eqref{eq:recipe-search}, the researcher chooses $\searchvar{r_t}$ and the environment carries out $\operatorname{Exec}$ and $\operatorname{Eval}$. The environment owns the tedious recurring work: resource allocation, job execution, checkpoint storage, evaluation, and recovery. This way the researcher's effort goes into the research itself. The evaluator runs outside the researcher's sandbox and grades by verifiable rules or programs, so its feedback follows fixed rules and stays beyond the reach of the researcher it measures.

\paragraph{Externalize the research state.}
The research record is the history $\historyvar{\mathcal H}$ of \eqref{eq:recipe-search}. It links each request to its code, data, and input checkpoints, and to the jobs, output checkpoints, and measurements that follow.
It holds four kinds of entries. The journal is a time-ordered log with one line for every accepted request and every change of state. The lineage records which checkpoints each checkpoint was made from. The transcript is the full text of every agent session. The workspace history is the version history of the workspace, including the researchers' notes. Because every checkpoint points to its parents, we can trace any score back to the base model, through every training stage and merge that produced it.  The record lives on storage outside the agent's context, so a researcher
can resume a session, start a fresh one, or be moved to a new node and still
have the same evidence. For the same reason the researchers can work as a
team: each reads its teammates' workspaces, results, and journals and can fetch
their saved checkpoints.

\paragraph{Keep the research loop productive.}
A training job can take hours, and each of its checkpoints can be evaluated only once it is saved: training and evaluation together make up the feedback delay $\feedbackvar{\tau_{\mathrm{fb}}}$ of \cref{sec:autoresearch}. A researcher that waited for each result before starting the next
experiment would leave most of its GPUs idle, so Athanor lets it keep several
experiments in flight.
Every job runs in the background and belongs to the environment rather than to
the agent's session: the agent can go on reading, planning, and starting other
jobs, and the job survives if the session ends. When a job or an evaluation
finishes, the services push its result into the agent's session, so one score
can inform the next decision while other jobs are still running. \Cref{app:system-details} gives the implementation of all three principles.

\section{AutoPostTrain at Scale}\label{sec:campaign}

We test Athanor in one campaign, in which a team of four agents post-trains Qwen3-4B-Base for 104 hours.

\subsection{Experimental Setting}\label{sec:setting}

\paragraph{Agents and goal.}
Four researcher agents, each Claude Opus 5.5 in Claude Code, work as one team;
we call them Agents 1--4. Each has one node, a sandbox,
read-only web access, and the \texttt{athanor} command; a node is a GPU server with eight GPUs, each with 80\,GB of HBM3 memory, and a high-bandwidth GPU interconnect. The goal they read is
one sentence: \textit{raise the development-suite score of a model post-trained from
Qwen3-4B-Base as high as possible}. It names no target score and prescribes no method. The agents
see the development scores of three reference models: the base, the \emph{default starter} (the starter code run with its default settings and no research), and Instruct-2507, so the level of the vendor's release is visible to them.

\paragraph{Data and protocol.}
The agents may train on public datasets, problems they generate, and samples
from checkpoints they have trained.
 They may not call or download any other model, and humans give no research advice. The selection rule is fixed before the campaign starts: the
development suite selects the team's submission, and the held-out suite,
scored after the campaign, gives the result. We count hours from the start of the research window, the period in which the agents work.  \Cref{app:campaign-protocol} gives the setting and timeline.

\paragraph{Evaluation suites.}
Each suite has six tasks, one per capability, and its score is their plain
mean. Each held-out task is the twin of one development task: it measures the same
capability on different material, taken from a different source, from a later
time window, or from task types the agents never saw. The pairs, development task first, are: mathematics contests from early 2025 and from November 2025 on~\citep{balunovic2026matharena}; two disjoint halves of ZebraLogic~\citep{lin2025zebralogic}; LiveCodeBench problems from 2024 and from 2025~\citep{jain2025livecodebench}; MMLU-Pro~\citep{wang2024mmlu} and SuperGPQA~\citep{du2026supergpqa}; IFEval~\citep{zhou2023instruction} and IFBench~\citep{pyatkin2026generalizing}; and, for multi-step action, four ALFWorld task types and the other two~\citep{shridhar2021alfworld} (\cref{tab:suites} gives the number of items in each). In this paper a \emph{seed} is one complete evaluation of a suite; seeds differ in the sampling and in the small run-to-run differences of batched inference, and $\pm$ gives the standard error of the suite mean over items.

\paragraph{References and measurement protocols.}
We compare against the base, the default starter, and the vendor's three post-trained Qwen3-4B releases: Qwen3-4B-Instruct-2507, the original hybrid Qwen3-4B, and Qwen3-4B-Thinking-2507; Qwen3-4B and Thinking-2507 write out their reasoning before they answer. All models are scored by the same frozen evaluator under the \emph{campaign protocol}: mathematics is sampled, the other tasks are decoded greedily, and an answer may use up to 16,384 output tokens on mathematics, logic, and code, 8,192 on knowledge and instruction following, and 256 per ALFWorld step. Under these limits the two reasoning releases often run out of tokens before they answer, so after the campaign we add a second measurement of the references and the submission with up to 32,768 output tokens (8,192 on instruction following and per ALFWorld step) and one sampling setting for every model, the setting Qwen recommends for Qwen3 in thinking mode (temperature 0.6, top-$p$ 0.95, top-$k$ 20); we call this the \emph{final measurement}. This shared setting does not handicap Instruct-2507: with its own recommended sampling and chat template it scores 66.4 and 47.5, no higher (\cref{app:campaign-protocol}).

\Needspace{8\baselineskip}
\subsection{Model Improvement}\label{sec:results}

\begin{table}[t]
 \centering
 \small
 \setlength{\tabcolsep}{3.8pt}
 \caption{\textbf{Scores of the team's submission and the references} on the development suite, which the agents optimize, and the held-out suite, which they never see. Mat, Log, Cod, Kno, Ins, and Act are mathematics, logic, code, knowledge, instruction following, and multi-step action (\cref{tab:suites}). Top: the campaign protocol; bottom: the final measurement (\cref{sec:setting}). Each mean is over two to four seeds, with a standard error over items of 0.5 to 1.4 points. Our submission's means are in bold; the final measurement does not include the default starter.}
 \label{tab:c1-main}
 \begin{tabular}{@{}lccccccc|ccccccc@{}}
 \toprule
 \tablehead
 & \multicolumn{7}{c|}{Development suite} & \multicolumn{7}{c}{Held-out suite} \\
 \tablehead
 Model & Mat & Log & Cod & Kno & Ins & Act & \textbf{Mean} & Mat & Log & Cod & Kno & Ins & Act & \textbf{Mean} \\
 \midrule
 \multicolumn{15}{@{}l}{\textit{Campaign protocol (up to 16,384 output tokens)}} \\
 Qwen3-4B-Base & 9.2 & 11.0 & 22.3 & 15.3 & 38.9 & 0.0 & 16.1 & 3.4 & 11.4 & 18.3 & 7.8 & 11.5 & 0.0 & 8.7 \\
 Default starter & 6.2 & 13.1 & 20.9 & 54.6 & 72.5 & 51.3 & 36.4 & 3.4 & 16.2 & 16.5 & 28.8 & 22.6 & 34.9 & 20.4 \\
 \keyrow Team submission (ours) & 37.2 & 63.4 & 47.1 & 65.1 & 84.8 & 98.9 & \textbf{66.1} & 29.7 & 63.7 & 34.9 & 36.0 & 37.0 & 78.8 & \textbf{46.7} \\
 Qwen3-4B-Instruct-2507 & 42.9 & 80.4 & 43.9 & 63.8 & 84.9 & 82.4 & 66.4 & 42.2 & 81.0 & 31.4 & 36.4 & 31.1 & 65.9 & 48.0 \\
 Qwen3-4B (hybrid) & 44.7 & 80.8 & 56.4 & 64.1 & 80.2 & 24.2 & 58.4 & 40.5 & 80.3 & 36.6 & 35.0 & 27.3 & 8.2 & 38.0 \\
 Qwen3-4B-Thinking-2507 & 32.7 & 89.8 & 63.3 & 68.7 & 86.1 & 0.0 & 56.8 & 31.0 & 89.1 & 41.7 & 39.0 & 44.8 & 0.0 & 41.0 \\
 \midrule
 \multicolumn{15}{@{}l}{\textit{Final measurement (up to 32,768 output tokens, sampled)}} \\
 Qwen3-4B-Base & 8.9 & 10.3 & 20.4 & 20.6 & 35.3 & 0.0 & 15.9 & 4.3 & 10.3 & 14.7 & 12.7 & 13.0 & 0.0 & 9.2 \\
 \keyrow Team submission (ours) & 38.4 & 63.0 & 51.1 & 70.4 & 85.2 & 95.6 & \textbf{67.3} & 33.3 & 62.5 & 37.9 & 43.3 & 37.6 & 74.7 & \textbf{48.2} \\
 Qwen3-4B-Instruct-2507 & 44.2 & 79.5 & 45.1 & 64.1 & 84.1 & 81.8 & 66.5 & 42.5 & 77.7 & 33.6 & 36.5 & 30.3 & 66.8 & 47.9 \\
 Qwen3-4B (hybrid) & 53.7 & 82.1 & 68.4 & 70.9 & 83.3 & 81.3 & 73.3 & 50.3 & 81.4 & 45.5 & 42.7 & 31.1 & 52.6 & 50.6 \\
 Qwen3-4B-Thinking-2507 & 62.7 & 92.5 & 74.7 & 74.4 & 85.8 & 93.1 & 80.5 & 60.5 & 92.0 & 55.1 & 49.6 & 46.5 & 74.1 & 63.0 \\
 \bottomrule
 \end{tabular}
\end{table}

\paragraph{Search from the base.}
We report the campaign protocol first, then the final measurement that gives the abstract's numbers. Under the campaign protocol (\cref{tab:c1-main}, top), in 104 hours the best single evaluation, the \emph{frontier} (the staircase of \cref{fig:c1-climb}; the best observed suite mean of $\obsvar{\widehat{\mathbf p}_t}$ in \eqref{eq:recipe-search}), rises from 16.1 to 66.9, above the 66.4 of Qwen3-4B-Instruct-2507. Without any research, the default starter reaches 36.4. A single evaluation of the best candidate tends to run high, because a search that keeps the best single evaluation selects on the evaluation noise $\feedbackvar{\xi_t}$. The agents see this in their own record and, from the second day, re-evaluate 23 of their best candidates on further seeds: for 19 of them the first seed scores higher than the mean of the later seeds (sign test, $p=0.003$), and across all 23 the first seed is higher by 0.4 points on average and by at most 1.1 (\cref{fig:c1-science}a). The reported result is therefore the team's submission, averaged over several seeds.  Each agent submits one checkpoint, and the team's submission is the one with the highest development score among the four, by a rule fixed before the start; this is \agent{3}'s checkpoint, an average of its own and \agent{2}'s self-distilled models:  $66.1\pm0.8$ averaged over three seeds, against 66.4 for Instruct-2507 (\cref{tab:c1-main}).  Because ideas and checkpoints spread through the team (\cref{sec:f-team}), all four agents arrive at this
level: the other three submissions are each within two standard errors of it (\cref{tab:sealed}).

\paragraph{Held-out transfer.}
The gain holds on tasks the agents never saw. Under the campaign protocol, on the held-out suite, the submission closes 97\% of the gap between the base and Instruct-2507, where the default starter, at 20.4, closes 30\%. The submission scores $46.7\pm1.3$ over two seeds, against 8.7 for the base and $48.0\pm1.4$ for Instruct-2507. Compared with
Instruct-2507 on the same items, the submission is 1.3 points lower, and the 95\% confidence interval for this
difference, paired over items, runs from $-4.0$ to $+1.3$: the two models are not distinguishable. All four agents' submissions reach this level: they score 45.9 to 47.2, and for each the 95\% interval for the difference from Instruct-2507 contains zero.

\paragraph{Task profile and comparison with reasoning releases.}
The team reaches the vendor's mean by a different route, and the difference
repeats on both suites; \cref{fig:c1-tasks} shows how each task climbs. The submission is above Instruct-2507 on multi-step
action (98.9 against 82.4 on development, 78.8 against 65.9 held-out) and code
(47.1 against 43.9, and 34.9 against 31.4), above or level on knowledge and instruction
following, and below on logic (63.4 against 80.4, and 63.7 against 81.0) and mathematics (37.2 against 42.9, and 29.7 against 42.2). The means are level because multi-step action offsets mathematics and logic: without it, the submission trails Instruct-2507 by 3.7 points on development and by 4.2 on the held-out suite. The agents fine-tune on ALFWorld trajectories, and the gain carries over to the two task types they never saw. The team's models, fine-tuned on reasoning traces, write out their reasoning before they answer, so on logic and mathematics many answers hit the output limit before they end. For the team's submission, 18\% of the logic answers and 24\% of the mathematics answers are cut off, against 1\% and 10\% for Instruct-2507 (\cref{app:plateau}).

The final measurement (\cref{tab:c1-main}, bottom) raises the team's means to
67.3 on development and 48.2 on the held-out suite, level with or above Instruct-2507 (66.5 and 47.9; \cref{tab:v4-tasks} gives all four submissions); the paired 95\% intervals of the differences, $-0.8$ to $+2.4$ and $-2.0$ to $+2.6$, contain zero. With longer output limits and sampled decoding, the submission's logic score does not rise (63.4 to 63.0, against 79.5 for Instruct-2507), so truncation is not the whole gap; most of the gap lies in the larger logic puzzles (\cref{app:plateau}). The vendor's two reasoning releases
score higher still (73.3 and 50.6 for the hybrid release, 80.5 and 63.0 for
Thinking-2507), mostly on mathematics, logic, and code: the team reaches the
vendor's instruct release but not yet its reasoning releases.

\subsection{Execution Effectiveness}\label{sec:execution}
The result above rests on the environment carrying out the agents' experiments. This section measures where the agents' time, the GPU hours, and the money go, and how the campaign runs over four days.

\begin{figure}[!t]
 \centering
 \includegraphics[width=\textwidth]{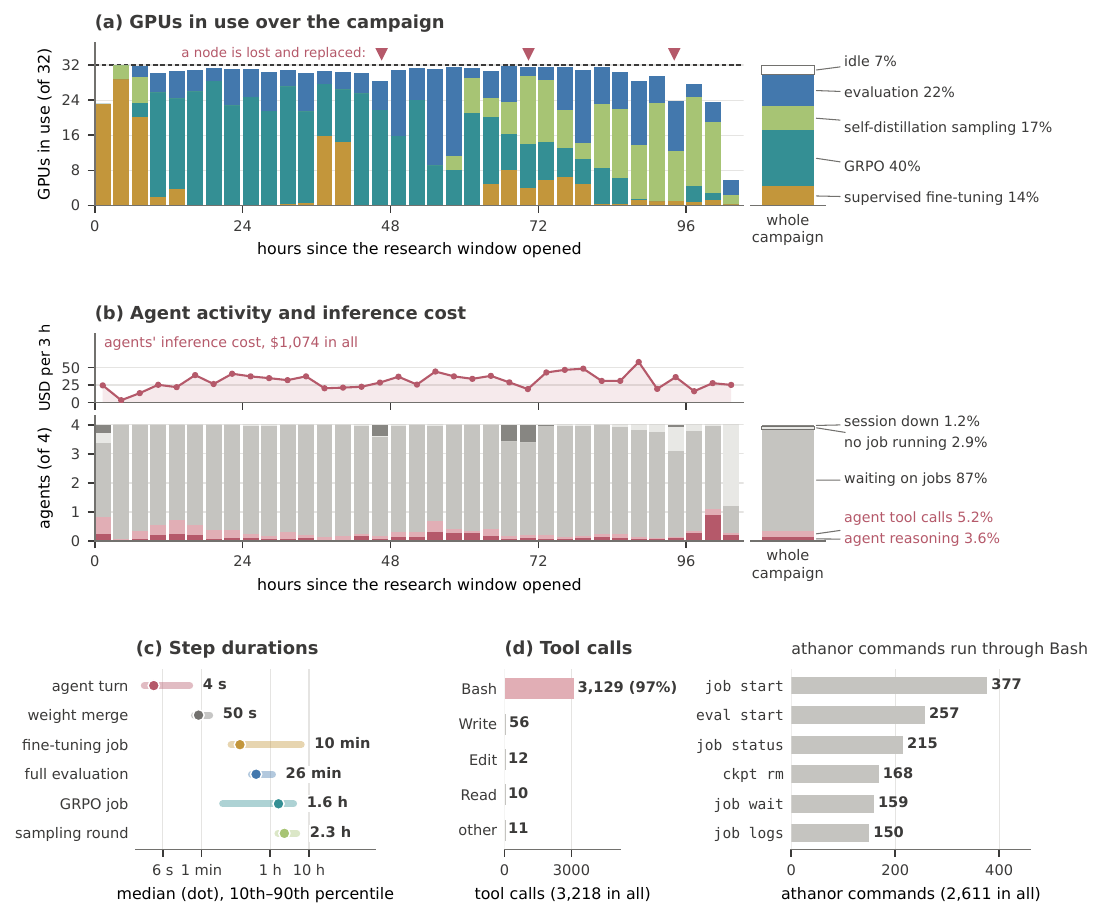}
 \caption{\textbf{The GPUs stay busy while the agents spend most of their time waiting on jobs.} (a) Jobs hold the GPUs for 93\% of the GPU hours, shown as three-hour means by phase of the recipe; triangles mark the three nodes the platform took back. (b) The agents reason for 3.6\% of the session time and wait on jobs for 87\%; above, their inference cost as Claude Code counts it comes to \$1,074. In (a) and (b) the column at the right gives the whole campaign; \cref{fig:c1-lanes} shows the GPUs and the turns of each agent. (c) An agent turn takes 4 seconds at the median, a full evaluation takes 26 minutes, and a GRPO job takes 1.6 hours; a sampling round draws the self-distillation samples (dot: median; bar: 10th to 90th percentile); most fine-tuning jobs are short, while each agent's first full fine-tuning takes 4.4 to 8.5 hours. (d) 97\% of the agents' tool calls run Bash, and the \texttt{athanor} commands they run most often start jobs and evaluations, check their state, and delete checkpoints to free disk space (\texttt{ckpt rm}).
 }
 \label{fig:c1-execution}
\end{figure}

\paragraph{Agent efficiency.}
The agents spend their time on the research, not on rebuilding the experiment infrastructure, and once a session has submitted a job, it waits for the result and spends no tokens while waiting (\cref{fig:c1-execution}b and d; \cref{fig:c1-lanes} shows each agent separately). Of the roughly 416 agent hours of the campaign, an agent is working (reasoning or running its tool calls) for about 9\%. The agents' reasoning, in which they read results, analyze them, and decide, takes 15 hours in all (3.6\%). Almost all of the agents' tool calls run shell commands, most often \texttt{athanor} commands that start jobs and evaluations and read their state; 41 of the 2,611 \texttt{athanor} commands fail, mostly on a wrong argument or name.
For 87\% of the time a session waits while its jobs run, until a result wakes the agent. A turn of agent work lasts 4 seconds at the median, while a GRPO job takes 1.6 hours and a full evaluation takes 26 minutes (\cref{fig:c1-execution}c); the GRPO job and the evaluation together make up the feedback delay $\feedbackvar{\tau_{\mathrm{fb}}}$. The GPUs therefore almost never wait for an agent: less than 1\% of the GPU hours pass while an idle GPU waits for its agent.

\paragraph{System efficiency.}
Jobs hold the GPUs for 93\% of the available GPU hours (3,086 of 3,315): of the available hours, 54\% go to training (40\% go to GRPO alone), 17\% go to sampling, almost all for self-distillation, and 22\% go to evaluation (\cref{fig:c1-execution}a). Holding a GPU is not computing on it: by the utilization the GPUs report, they compute for about 75\% of the available hours. Part of the gap lies inside the training and evaluation stacks: a GRPO step waits for the longest rollout of its batch, which leaves GPUs below 10\% utilization for 155 GPU hours, and jobs and evaluations spend about 90 GPU hours starting up before their first step.

\paragraph{Cost.}
The campaign has two costs, the GPUs and the agents' inference. Claude Code itself counts what each turn costs at public list prices: \$1,074 for the whole team.
To put the two resources on one scale, we price the GPU time at \$3.49 per GPU
hour, the median on-demand rental price of a GPU of this class across 41 cloud providers at the time of writing.\footnote{GetDeploying, week of October 5, 2026: \url{https://getdeploying.com/gpu-price-trends}.} At that price the 3,086 GPU hours that jobs hold cost about \$10,800 and the campaign costs about \$11,800 in all, of which the agents' own
inference is 9\% (\cref{fig:c1-cost}a). The frontier reaches about 50 after \$1,100 and about 60 after \$2,400, and the submission is first evaluated after about \$10,300 (\cref{fig:c1-cost}b); the last six points, from about 60 to the submission's 66.1, cost about \$8,000.

\begin{figure}[t]
 \centering
 \includegraphics[width=\textwidth]{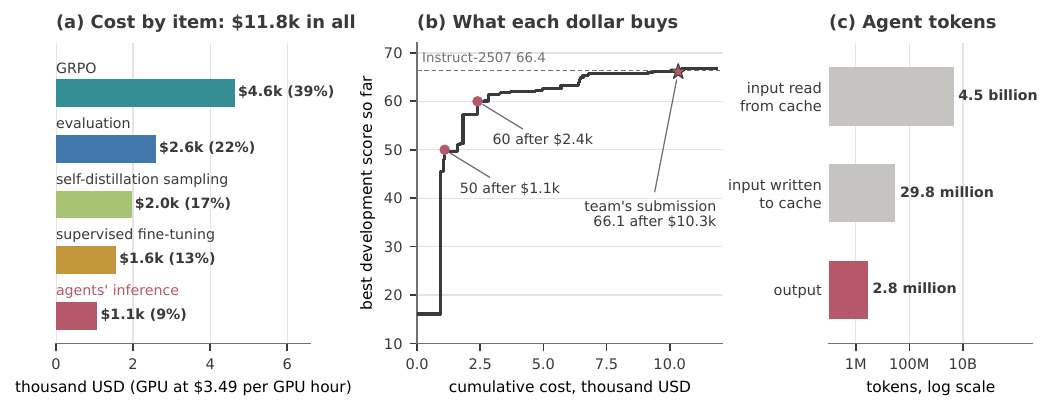}
 \caption{\textbf{The campaign costs about \$11,800, most of it GPU time.} (a) The GPU hours cost \$10,800 at \$3.49 per GPU hour, and the agents' inference, as Claude Code counts it, is 9\% of the total. (b) The best development score so far rises in steps; the curve follows single evaluations, and the star marks the first evaluation of the team's submission, after \$10,300, and its label gives the three-seed mean. (c) Almost all input of the agents' 8,243 model calls is read from the prompt cache: 4.5 billion tokens, against 29.8 million input tokens written to the cache.}
 \label{fig:c1-cost}
\end{figure}

\paragraph{Long-horizon reliability.}
The campaign runs for 104 hours, and the environment absorbs what goes wrong in that time. The platform takes back three of the four nodes during the run, at hours 46, 70, and 94. Each outage lasts 11 to 69 minutes and is added to that agent's window; the session resumes on a new node from its record, and what is lost is the jobs and evaluations running at that moment, 23 GPU hours of work in all, together with any files kept only on the node's local disk.
Requests rarely fail: 26 of 500 jobs end failed, most of them on errors in the agents' own programs, and 2 evaluations end failed. After the campaign every submitted checkpoint, its evaluations, and its lineage are recovered from the record.

\section{Trajectory Analysis}\label{sec:discoveries}

Because Athanor records every request, result, and checkpoint lineage, the campaign can be traced step by step from the base to the submission. The record of the campaign holds 500 jobs, 255 requested full evaluations (each one seed of the whole development suite; together they score 174 checkpoints), 57 checkpoint fetches between agents, and the transcripts and
notes of the four agents. We summarize it as four findings (\cref{tab:findings}): the first describes the recipe the team found, and the other three describe how the team searched. The agents' own notes behind each case are in \cref{app:trajectories}.

\begin{figure}[t]
 \centering
 \includegraphics[width=\textwidth]{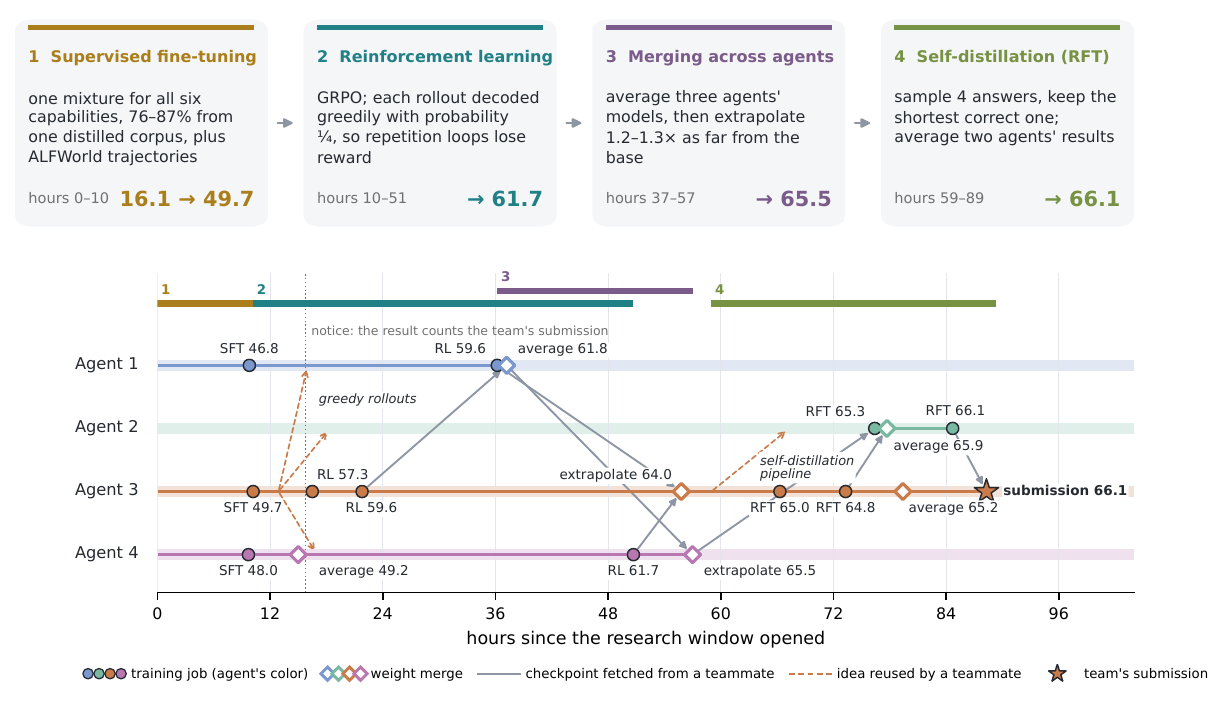}
 \caption{\textbf{The team's recipe: four methods, composed by the agents.} Top: one card per step, with what the agents did and the best development score among the submission's ancestors at that step; these ancestors come from different agents (49.7 is \agent{3}'s fine-tuned model, 61.7 \agent{4}'s GRPO model). Bottom: the same lineage on a timeline, one lane per agent, with the hours of each step marked in its card's color. Circles are training jobs, diamonds are weight merges, and every score is a mean over its seeds. The star is the submission, the average of \agent{2}'s second self-distilled model and \agent{3}'s average of its two rounds. An edge across lanes is a checkpoint fetched from a teammate, and the edges into a diamond are the models it averages (\agent{4}'s first diamond averages two of its own fine-tuned models). A dashed arrow is an idea reused by a teammate. Some intermediate checkpoints are not drawn; \cref{fig:c1-search} shows them all. The dotted line marks hour 15.8, when the agents learn that the result counts the team's submission. \Cref{tab:lineage-steps} lists the training and merge steps.}
 \label{fig:c1-recipe}

\end{figure}

\begin{table}[t]
 \centering
 \footnotesize
 \setlength{\tabcolsep}{4pt}
 \caption{\textbf{Four findings from the record of the campaign}, with the evidence for each.}
 \label{tab:findings}
 \begin{tabularx}{\linewidth}{@{}l >{\raggedright\arraybackslash\hsize=0.9\hsize}X >{\raggedright\arraybackslash\hsize=1.1\hsize}X@{}}
 \toprule
 \tablehead
 Finding & What the record shows & Key evidence \\
 \midrule
 (1) Recombination & Every step from the base to the team's submission is a known method; the agents choose which methods to combine, in what order, and from whose checkpoint & Among the submission's ancestors, fine-tuning reaches 49.7, GRPO with a quarter of the rollouts greedy 61.7, merges and extrapolation 65.5, and self-distillation 66.1 (\cref{fig:c1-recipe}); the agents write the rollout change, the merge recipes, and the self-distillation programs \\
    \midrule
 (2) Team progress & Ideas and checkpoints move between agents within hours, and the best checkpoints combine several agents' training & 57 checkpoint fetches from teammates and 160 versions of the shared notes (\cref{fig:c1-team}a); one agent's training alone reaches 61.7, merges across agents 65.5; after hour 21 every new best combines several agents \\
    \midrule
 (3) Recipe convergence & All four agents build on the same corpus within the first hour; the alternatives they try later score lower and are dropped & AM-Thinking-v1-Distilled is 76--87\% of the public-data rows of every agent's first mixture (\cref{fig:c1-team}c); two alternative lineages are each dropped after one round of experiments; no experiment compares two data sources under otherwise identical conditions \\
    \midrule
 (4) Uneven data sense & The agents catch problems of format, length, scorability, and difficulty by inspecting rows, and rarely verify that answers are correct & 30 to 70 row excerpts read per agent in the first six hours; 1,093 rows that cannot be scored dropped; a planned filter for answers that look like noise never run \\
 \bottomrule
 \end{tabularx}
\end{table}

\subsection{Recombination}\label{sec:f-recombination}
The recipe that reaches the vendor's instruct release is built from known methods; the agents supply the composition and the fixes that make it work.
\Cref{fig:c1-recipe} shows the lineage of the team's submission, with the score each stage reaches. It contains
training by all four agents and passes through four stages, each a known
method. \textit{Supervised fine-tuning}: within the first hour,
each agent builds its own mixture of about 300 to 600 thousand rows covering all six
capabilities; it then fine-tunes the base on it for one epoch. \textit{Reinforcement learning}: the agents continue with
GRPO~\citep{shao2024deepseekmath} on rule-based rewards. \agent{3} finds that its fine-tuned model falls into repetition loops when it decodes greedily, as the evaluator does on five of the six tasks, and that GRPO does not fix this, because it trains on sampled answers, which rarely loop. \agent{3} therefore decodes each GRPO rollout greedily with probability one quarter; with this change and a rebuilt set of training problems, its next checkpoint rises from 49.7 to 57.3 (\cref{tr:main}). \textit{Merging}: the agents average models from different agents~\citep{wortsman2022soups} and then extrapolate. \agent{4} takes a weighted average of three agents' models, and \agent{3} scales its difference from the base, its task vector~\citep{ilharco2023editing}, by a factor $f$ of 1.2 to 1.3, trading some knowledge for logic (\cref{fig:c1-science}b; \cref{app:posttrain-science}).
\textit{Self-distillation}: the agents sample four answers per training problem
from the merged model, keep the shortest answer that the rule-based reward marks correct, and
fine-tune on the result, a form of rejection-sampling fine-tuning
(RFT)~\citep{zelikman2022star,yuan2023scaling}. Two agents each run self-distillation rounds, and averaging their results gives the submission.

Of these four methods, the starter code provides fine-tuning, GRPO, and merging~\citep{mergekit,yadav2023ties,yu2024dare}. The agents choose and write
everything else themselves: the data mixtures, the rollout change, the merge
recipes and their weights, and the self-distillation programs.

\Cref{tr:main} follows the rollout change through one cycle of the loop of
\cref{alg:autoresearch}, from the measurement to the hand-over to the team.
The bold headings are our summaries. The lines marked \textsc{Researcher} are copied verbatim from \agent{3}'s notes. In them, r1--r4 are Agents 1--4; zebra, LCB, mmlu (or mcq), and ifeval are the
logic, code, knowledge, and instruction-following tasks; ``budget'' is the output limit; T is the sampling temperature; sN is step N; and ``val'' is the training job's own small validation set, decoded greedily and scored from 0 to 1. \Cref{app:episode} shows the agent's turn that writes the change.

\begin{agenttrace}[breakable=false,label={tr:main}]{From a diagnosis to a method the whole team uses (\agent{3}, hours 10--22)}
\small
\textbf{Hour 10: greedy answers fall into loops.} \agent{3}'s first fine-tuned model scores 49.67 on the development suite. The agent examines every answer of this evaluation that was cut off at the output limit. For each task it records three numbers: the share of answers that were cut off, how many of the cut answers are repetition loops, and the accuracy of the answers that finished.

\smallskip
\textcolor{athanorblue}{\textsc{Researcher}}\quad\emph{Truncated at budget /
loops among them / acc of finished: math 66\,\% / 46 of 486 / 76\,\%; zebra
58\,\% / 238 of 292 / 71\,\%; LCB 69\,\% / 331 of 351 / 90\,\% \ldots{} Greedy
loops are the main loss on zebra/LCB/mmlu/ifeval; length on math.}

\smallskip
\textbf{Hours 10\textendash 13: GRPO on sampled answers does not fix the loops.} The agent starts GRPO and reads
its own rollouts while the job runs.

\smallskip
\textcolor{athanorblue}{\textsc{Researcher}}\quad\emph{grpo1 step-25 greedy val
vs step 0: ifeval .31->.30, zebra .08->.09, math .36->.23, code .19->.22, mmlu
.41->.42; greedy truncation unchanged \ldots{} Sampled rollouts (T .8) rarely
loop, so GRPO on them does not touch greedy loops. Stopped grpo1 at step 25.}

\smallskip
\textbf{Hours 12\textendash 13: the fix, a quarter of the rollouts greedy.} The agent writes a rollout loop that decodes each answer greedily with probability one quarter. A greedy answer that falls into a loop scores zero, so whenever a sampled answer to the same problem scores higher, the looping answer gets a negative advantage. The agent then restarts training with the new loop
on a rebuilt set of problems.

\smallskip
\textbf{Hour 15: on the validation answers, loops drop from 86 to 18.}

\smallskip
\textcolor{athanorblue}{\textsc{Researcher}}\quad\emph{grpo2 step-20 greedy val
(vs step 0): math .33->.53, code .22->.34, mcq .44->.55, ifeval .27->.36, zebra
.08->.14; truncations 157->95, loops 86->18. Greedy-mix rollouts work. \ldots{} plan full eval at step 40.}

\smallskip
\textbf{Hours 16\textendash 22: 49.7 becomes 57.3, and the team adopts the fix.} The full evaluation at step 40 gives
57.27. \agent{3} writes the diagnosis and the code into its shared notes
(\emph{``The SFT models' main loss is greedy repetition loops, and GRPO on
sampled rollouts does not fix them \ldots{} Fix that works: make a share of the
GRPO rollouts greedy''}). Within hours the three teammates copy the loop;
\agent{1} reports \emph{``v3 s50 (52.18) + 25 steps with r3's 25\% greedy
rollouts \ldots{} 56.50 \ldots{} Thanks r3''}.
\end{agenttrace}

\subsection{Team Progress}\label{sec:f-team}
Ideas and checkpoints move between agents within hours, and the team's best
checkpoints combine the training of several agents. Before hour 15.8 the agents read each other's files 37 times, but they write no notes for their
teammates and fetch no checkpoints. At hour 15.8 the operator tells the agents that the final result counts the team's submission, a rule fixed before the start that their documents had not stated; we call this the \emph{notice}.  Within minutes each agent writes its first note
for its teammates, and the first checkpoint is fetched from a teammate 56
seconds after the notice arrives (\cref{fig:c1-team}a).
From then on the agents share without further prompting. All four agents are
using \agent{3}'s rollout change within six hours of its first use, and \agent{1}'s and \agent{4}'s
notes credit \agent{3} for it. \agent{4} takes up \agent{3}'s extrapolated merge within hours. \agent{2} copies \agent{3}'s self-distillation programs on the day they are written.

\begin{figure}[H]
 \centering
 \includegraphics[width=\textwidth]{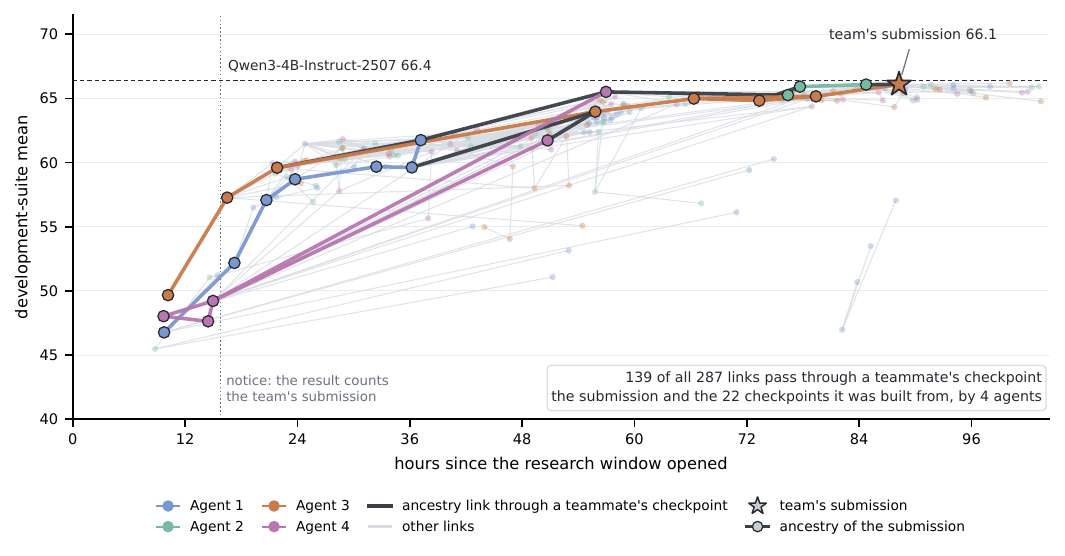}
 \caption{\textbf{The whole search.} Each dot is one of the 174 fully evaluated checkpoints, at the hour of its first evaluation and its mean development score, colored by agent; each line joins a checkpoint to the evaluated ones it was made from. The team's submission (star) and the 22 evaluated checkpoints it was built from are drawn in full, and their links are dark where the path passes through a checkpoint fetched from a teammate. The base (16.1) is off the axis.}
 \label{fig:c1-search}
\end{figure}

From hour 21, when \agent{4}'s average of two teammates' models (59.95, \cref{tr:soup}) becomes the best checkpoint, every checkpoint that sets a new best score is either a merge
of models from several agents or was trained from such a merge. We link each
evaluated checkpoint to the nearest evaluated checkpoints it was made from.
This gives 287 links, and for 139 of them the path between the two checkpoints includes a checkpoint
fetched from a teammate (\cref{fig:c1-search}; \cref{app:team} lists a selection of the
exchanges). The agents' lineages also grow together over time.
Before the notice, no evaluated checkpoint contains any training done by a
teammate. Of the 32 checkpoints evaluated after hour 84, 28 are built on training by all four agents (\cref{fig:c1-team}b). This fits the finding of Park et al.\ that sharing helps when agents have a verifier to tell which teammates' results are worth building on~\citep{park2026communication}; here the development suite plays this role by scoring every checkpoint.

\begin{figure}[t]
 \centering
 \includegraphics[width=\textwidth]{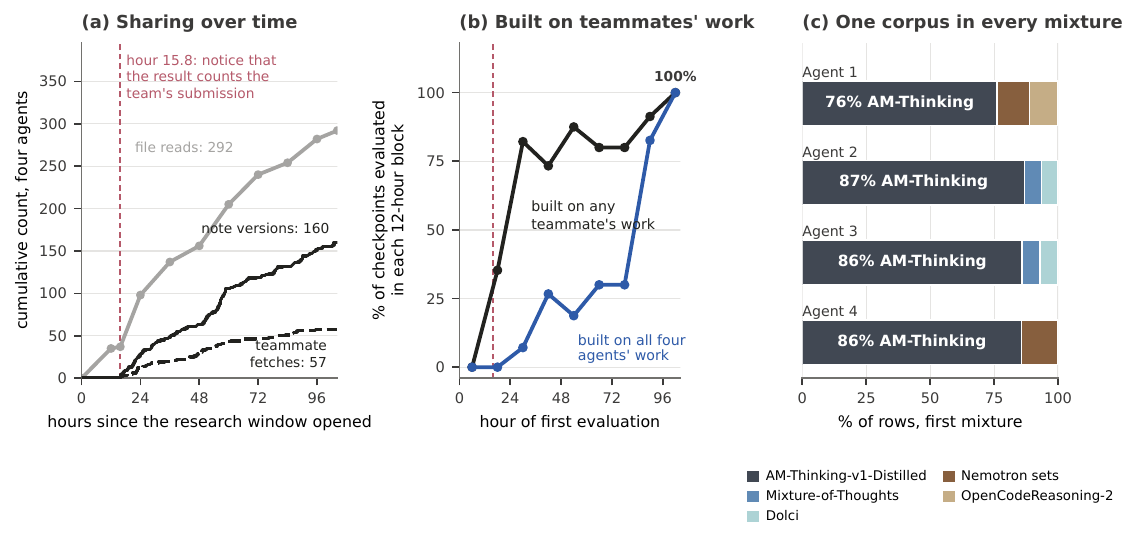}
 \caption{\textbf{The team starts sharing at the notice and ends up building on all four agents' work.} (a) Before the notice at hour 15.8 the agents only read teammates' files; after it they write 160 versions of the shared notes and fetch 57 checkpoints from teammates. Reads are known only as totals per twelve-hour block and at the notice (dots). (b) The share of newly evaluated checkpoints that contain a teammate's training (black) jumps after the notice, and in the last block, from hour 96, every new checkpoint contains the training of all four agents (blue). (c) AM-Thinking-v1-Distilled makes up 76--87\% of every agent's first fine-tuning mixture, as a share of its rows from public datasets.
 }
 \label{fig:c1-team}
\end{figure}

\subsection{Recipe Convergence}\label{sec:f-convergence}
The four agents converge on one recipe early. Within the first hour, each agent chooses
AM-Thinking-v1-Distilled~\citep{amthinking2025}, a public corpus of long reasoning traces distilled from a stronger model, as the largest source of its fine-tuning data, and each adds
a few other sources. No agent runs a controlled comparison of this corpus against another source. The corpus makes up 76--87\% of the public-data rows of
each agent's first mixture (\cref{fig:c1-team}c). Their search over data stays narrow: of the 447 datasets in the pool, the agents look at 43 and use 27, and all four use eight of them (\cref{app:campaign-protocol}).

The team's attempts to diversify fail. At hour 36, \agent{3} notes that every fine-tuned model of the team is built on
that one corpus. To make the merges more diverse, it trains a new lineage from the base on other data. The new lineage stays near 55
while the team's best is near 62, and merging it into the main lineage lowers
the logic score, so \agent{3} drops it; a second alternative lineage, trained on shorter data,
also scores lower and is dropped. By the fourth
day, all four agents are refining one model and merging each other's variants
of it. Anthropic's study of multi-agent systems describes the same tendency: agents that
share a model, a harness, and a context take very similar actions, so one
agent's decision becomes the decision of all of
them~\citep{anthropic2026multiagent}.

\subsection{Uneven Data Sense}\label{sec:f-data}
The agents inspect their data and act on what they see. Each reads 30 to 70
excerpts of rows from candidate corpora in the first six hours, and later they
read their models' answers to judge the data behind them. They reject corpora whose reasoning traces exceed the output limit and drop 1,093 instruction rows that the reward function cannot score. They fix a deduplication step that had marked every puzzle as a duplicate and rebuild a puzzle pool that had been mostly small grids. After seeing that a model trained on puzzle-solver traces ``asserts unjustified deductions,'' they discard those traces. What they rarely do is check whether the
answers in the data are correct: one agent notes that some ground-truth answers
look ``like noise,'' plans a filter to remove them, and never runs it
(\cref{app:data-sense}).

\subsection{Why the Gains Slow}\label{sec:f-slow}
\Cref{fig:c1-climb} is the capability curve of autoresearch scaling (\cref{sec:autoresearch}): because jobs hold the GPUs for 93\% of the available hours, hours track compute (\cref{fig:c1-cost}b plots it against cost). The frontier reaches 60 within the first day; the remaining three days add about seven points, the largest share on logic, where the leading checkpoint's score rises from 49 to 65 (\cref{fig:c1-tasks}). The findings above suggest why. In the last two days no step on the submission's lineage brings in new training data: the gains of those days come from merging the team's models and from distilling them on their own samples (\cref{sec:f-recombination}). All of these models descend from one choice of data made in the first hour, so each merge combines closely related models, and the two lineages trained on other data score lower and are dropped (\cref{sec:f-convergence}).

\section{Insights and Discussion}\label{sec:insights}

\begin{figure}[t]
  \centering
  \includegraphics[width=\textwidth]{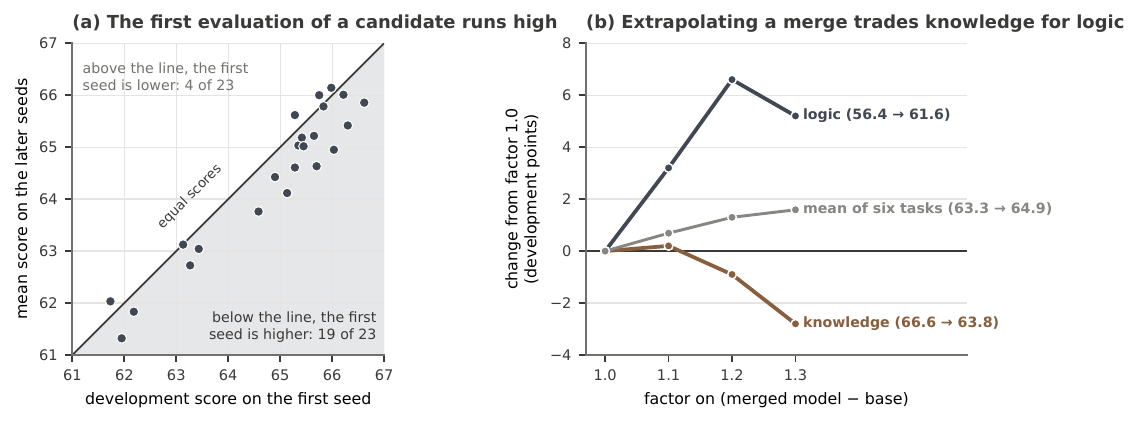}
  \caption{\textbf{Two observations from the record.} (a) For the 23 checkpoints evaluated on more than one seed, the first seed against the mean of the later ones: 19 lie below the diagonal, so a single evaluation tends to read high. (b) The change of the development mean and of two tasks, relative to $f=1.0$, when \agent{4}'s weighted average of three agents' models (63.3 at $f=1.0$; not drawn in \cref{fig:c1-recipe}) is moved by \agent{3} to base $+\,f\times$(merged model $-$ base) for $f$ from 1.0 to 1.3 (the 64.0 of \cref{fig:c1-recipe} is the four-seed mean at $f=1.2$), with the scores at the two ends in parentheses. Each point is a single seed, so differences under about one point may be noise.}
  \label{fig:c1-science}
\end{figure}

\paragraph{Automated scientific discovery of post-training.}
The record does more than document one result: it turns open questions of
post-training into measurements, each tied to the recipe and checkpoints that
produced it. If such campaigns run at scale, the first systematic understanding of how post-training shapes a model may come from AI agents studying AI rather than from people. In single experiments of this campaign, averaging two equally strong models from different lineages gains
two to three points, moving a merge further from the base trades knowledge for
logic (\cref{fig:c1-science}b), and \agent{3}'s first round of self-distillation adds about one point
(\cref{app:posttrain-science}). The agents draw these lessons themselves and
write them into their notes as rules for their teammates, such as \agent{4}'s
\emph{``averaging two EQUAL-strength RL models from different lineages gives
$\sim$+2.7''}. A search of this kind runs such experiments at the rate
the compute allows and keeps the evidence behind each conclusion. It can do so only because of what the record keeps. Post-training feedback is slow and noisy (\cref{sec:autoresearch}), so the cause of a score has to be recoverable afterwards, and Athanor keeps the answer to every item, repeated evaluations, the lineage of each model, and the teammates' work. The rollout change of \cref{sec:f-recombination} came from reading the scored answers, which show that many answers reaching the output limit are caught in a repetition loop; a record of scores alone would have hidden the cause.

\paragraph{Toward a broader autoresearch loop.}
The campaign fixes the suites and excludes teacher models; these boundaries keep the result checkable
and also limit what the agents can try. Everything else this campaign rests on was built by people, and each piece could be handed to agents, making the campaign one part of a larger loop in which AI improves AI research. The benchmark: agents could construct the evaluation suites for the capabilities they target. The environment: agents could build the next autoresearch environment, for post-training or for other domains, as we build Athanor through the meta loop of \cref{sec:athanor-philosophy}. The data: agents could run a data flywheel, generating, verifying, and cleaning their own training problems, so that synthetic data becomes an essential part of the search. And the researchers themselves: every campaign leaves a complete record of research at scale, with each decision tied to the evidence the agent saw and the result that followed, and such trajectories are training data for models built to do autoresearch. Each step that hands the agents more of the loop needs its own checks, against contamination and against agents that learn to please the evaluator.

\section{Related Work}\label{sec:related_work}

\paragraph{Automated post-training.}
PostTrainBench~\citep{ptb} has an agent post-train a small base model for one
benchmark on one GPU in ten hours, and Locus~\citep{locus} extends this
setting to parallel, multi-day search; both produce one specialist per
benchmark. RSI-Master and RSIGym also train per task, with access to external model APIs and with training and evaluation behind remote services, respectively~\citep{du2026rsimaster,meng2026rsigym}. RSI Arena~\citep{bakeai2026rsiarena} runs ten agents as competitors, each post-training the same 30B base model on a shared cluster, and ranks their models by human preference. Other benchmarks isolate fine-tuning or reinforcement-learning
pipelines~\citep{ftdojo,agent2rlbench}. An analysis of PostTrainBench
trajectories finds that agents rarely change strategy between consecutive
training attempts~\citep{lim2026missing}, a narrow search like the one we
observe. Our setting differs in that a team of agents works for days on one
generalist checkpoint, without a teacher model, and is scored on a held-out
twin suite.

\paragraph{Agents that improve AI systems.}
Agents now improve the systems around models: their own code and harness~\citep{zhang2026darwin,liu2026solpi,xia2026rrsi} and inference engineering~\citep{lmsys2026inferforge}, up to the production inference service of GLM's own models~\citep{zai2026glm}.  All of these
work on the layers around the model; our agents change the model's weights.

\paragraph{Autonomous research and teams of agents.}
Research agents now carry out experiments over hours to
days~\citep{aiscientist,alphaevolve,hambardzumyan2026aira2}, and benchmarks
score machine-learning engineering and research
tasks~\citep{mlebench,rebench,mlsbench,xu2026autolab}.
Work on teams of agents finds that identical agents sharing a workspace outperform independent ones when a verifier tells them whose progress to build on~\citep{park2026communication}, that agents sharing a model, a harness, and a context behave alike~\citep{anthropic2026multiagent}, and that a single agent can get stuck in an idea basin, a small region of ideas it keeps returning to~\citep{cai2026forkflush}; other systems share work among agents through a forum, a git history, or co-evolution~\citep{chen2026automatedalignment,zhang2026agora,qu2026coral}. Our team's convergence on one corpus resembles an idea basin shared by several agents, measured here on real training compute; \cref{app:related-tables} compares Athanor with neighboring systems.

\section{Limitations}\label{sec:limitations}

\paragraph{One campaign, one base model, and one agent model.}
The result comes from one campaign, with one team of four identical agents and one 4B base model. Its four submissions reach the same level, but they share their work and are not independent repeats; a repeat would show how much the result varies from campaign to campaign, and a campaign without sharing would show how much of it the team adds. We chose a 4B base because a model of this size can be trained and evaluated many times within one campaign; whether the recipe the team found carries over to larger models or to other model families is untested. The search itself may also change with the agents: agents built on another model may search differently, and a team that mixes models may converge less on one recipe than our team of identical agents did.

\paragraph{Not yet at the vendor's ceiling.}
The team reaches the vendor's instruct release, not its two reasoning releases, which score higher in the final measurement (\cref{tab:c1-main}). And although the agents call no teacher model, much of the public data they fine-tune on was distilled from stronger models by its publishers~\citep{amthinking2025,tulu3,nemotron}, so the result shows what agents can build from public resources, not from the base model alone.

\section{Conclusion}\label{sec:conclusion}

Athanor lets frontier agents run \apt{} end to end. In 104 hours, a team of
four agents takes Qwen3-4B-Base from 15.9 to 67.3 on six tasks and from 9.2 to
48.2 on a held-out twin suite, level with or above the vendor's instruct
release (66.5 and 47.9), by composing known methods and fixing the problems
their own measurements expose. The agents reach this level from public data, for about \$12,000 in GPU time and inference. Once the friction of execution is removed, the
clearest limit we observe is the diversity of their ideas.
This points to a future in which anyone who can score the behavior they want
automatically, and who has tokens and GPUs, can have agents post-train a
model for it. We hope that Athanor and our findings will provide a foundation
for scalable autonomous AI research.

\ifdefined\isarxiv\else %
\subsection*{AI Use Statement}
Following the ICLR 2027 policy on AI use, we disclose the following uses.
\textit{As the object of study:} in the reported campaign, the researcher
agents (Claude Opus 5.5) proposed and refined hypotheses, designed experiments,
implemented data-processing and training programs, and interpreted results;
these actions are what the paper analyzes. \textit{In building the system:}
coding agents helped write and revise Athanor's software, which the authors
designed, reviewed, and tested. \textit{In analysis and writing:} AI
assistants helped write the scripts that compute statistics from the
campaign records, create and refine figures, search and summarize related
literature, check and format references, and draft and edit text for
readability. The idea of autoresearch on post-training, the system design,
and the design of the campaigns are the authors' own. We have checked the
AI-assisted analyses, figures, references, and text against the underlying
records and sources, and we take full responsibility for the paper.
\fi

\ifdefined\isarxiv

\bibliographystyle{plainnat}
\begingroup\raggedright\setlength{\bibsep}{3pt plus 0.3ex}
\bibliography{ref}
\endgroup

\else

\bibliography{ref}
\bibliographystyle{iclr2027_conference}

\fi

\newpage
\onecolumn
\appendix

{\hypersetup{linkcolor=black}
\setlength{\parskip}{0pt}
\setcounter{tocdepth}{2}
\tableofcontents
}
\newpage

\begin{center}
	\textbf{\LARGE Appendix }
\end{center}

\section{System Details}\label{app:system-details}

This appendix gives the implementation of the three principles of \cref{sec:athanor-provides}: \emph{Execution and protected evaluation} implements the first; \emph{Workspaces and persistent state} and \emph{Research record} implement the second; and \emph{Background jobs and results} implements the third. \Cref{tab:roles} makes the roles of \cref{sec:athanor-philosophy} concrete: who decides each matter in a campaign and who carries it out.
\begin{table}[h]
  \centering
  \small
  \caption{\textbf{Who decides and who carries it out.}}
  \label{tab:roles}
  \begin{tabularx}{\linewidth}{@{}>{\raggedright\arraybackslash\hsize=1.25\hsize}X >{\raggedright\arraybackslash\hsize=0.8\hsize}X >{\raggedright\arraybackslash\hsize=0.95\hsize}X@{}}
    \toprule
    \tablehead
    Matter & Decided by & Carried out by \\
    \midrule
    Data, method, hyperparameters, merges, when to evaluate, which checkpoint to submit & Researcher & The environment's services, which put the request in a queue, assign GPUs to the job, and run it \\
    \midrule
    How the result is measured: which tasks are used, how answers are graded, the output limit, how many seeds the result uses, and which submission gives the result & Operator, declared before the start (the final measurement was added after the campaign; \cref{app:campaign-protocol}) & The evaluator and the services, as declared \\
    \midrule
    Nodes, the research window, and its extensions & Operator & Operator commands, each leaving a record \\
    \midrule
    The held-out suite & Operator & The operator, through a sealed evaluation that the researchers cannot see and are not told about \\
    \midrule
    The research record & No one decides this; the environment writes the record automatically & The environment; every analysis in this paper is derived from it \\
    \bottomrule
  \end{tabularx}
\end{table}

\paragraph{Execution and protected evaluation.}
Each researcher runs in a sandbox on its node, with its workspace, a scratch
disk, the base model, the data pool, and read-only views of its teammates'
workspaces and results. A service outside the sandbox accepts the requests of
the \texttt{athanor} command, queues them, binds GPUs, starts and stops jobs,
and accounts for GPU time. Evaluation runs outside the sandbox in its own
network namespace with a frozen copy of the evaluator, and only the scores and
scored answers of the development suite are copied back to the researcher.
Decontamination also runs outside the sandbox, and the researcher sees only how many rows it removed: a row of a training dataset is removed if it overlaps a benchmark item, either a development item or a held-out item. To test this, we lowercase the text and remove punctuation and markup. We then list all 13-word sequences in the row and in the item. The row is removed when at least half of the item's 13-word sequences also appear in the row. An item shorter than 13 words has no such sequence and is not matched by this rule.

\paragraph{Workspaces and persistent state.}
Everything the record needs is saved to durable storage as it is written: the workspace with its git history, the agent's home directory, the journal, the results of every evaluation and job, and every checkpoint saved as an artifact. Large files that the agents leave on the node's scratch disk, such as training datasets, are not. When the platform reclaims a node, the operator starts the session again on a new node. Requests that were still running on the lost node are marked as ended in the journal. The outage is added to the research window, and the agent session continues from its saved transcript and notes.

\paragraph{Research record.}
A request's journal line is written first, before the request starts anything. If the node fails, the journal still shows what was requested. For every checkpoint the record
keeps the job that wrote it, that job's inputs (datasets, parent checkpoints, and
the commit of the workspace when the job started), and the evaluations and
artifacts made from it, so the lineage of any checkpoint can be read back to the
base model. The researchers' notes are ordinary files in their workspaces.

\paragraph{Background jobs and results.}
Each job runs under the service, not under the agent's session, so it keeps running when the session ends. When a job or an evaluation changes state, its journal line is pushed into the agent's session as a message. Agents can also add their own watches, such as a program that wakes the agent when a given line appears in a training log.

\section{Feature Comparison with Neighboring Systems}\label{app:related-tables}
This appendix places the setting of \cref{sec:intro,sec:related_work} among neighboring systems.

\Cref{tab:closest-system-features} compares Athanor with two families of
systems. Post-training systems (Locus, AutoTrainess, A-Evolve-Training, the
automated alignment researchers (AAR), and RSIGym) also optimize model weights,
but none fully targets a single generalist model or records in full which
checkpoint was made from which. Research systems (AIRA$_2$, CORAL, Arbor,
Praxist) keep the state of the research, but they mostly improve a program or
the solution to one task rather than model weights. In the six feature
columns (all but Internet access), Athanor is the only system with full support (\fullmark) throughout.

\Cref{tab:automated-posttraining} compares what automated post-training
systems search over; they differ in what the agent controls.

\begin{table}[H]
  \centering
  \small
  \setlength{\tabcolsep}{4pt}
  \renewcommand{\arraystretch}{1.2}
  \caption{\textbf{Research and execution features of neighboring systems.}
  Internet access concerns the researcher's ability to retrieve external
  information or data during a campaign.
  }
  \label{tab:closest-system-features}
  \begin{tabularx}{\textwidth}{@{}>{\raggedright\arraybackslash}p{0.20\textwidth} *{7}{>{\centering\arraybackslash}X}@{}}
    \toprule
    \tablehead
    System & Internet access & Model as artifact & Generalist checkpoint & Checkpoint lineage & Async execution & Sealed final eval. & Research lineage \\
    \midrule
    Locus~\citep{locus} & Yes & \fullmark & \nomark & \partmark & \fullmark & \partmark & \partmark \\
    AutoTrainess~\citep{yu2026autotrainess} & Yes & \fullmark & \nomark & \partmark & \nomark & \partmark & \partmark \\
    A-Evolve-Training~\citep{shi2026evolve} & Yes$^{*}$ & \fullmark & \nomark & \nomark & \partmark & \partmark & \partmark \\
    AAR~\citep{chen2026automatedalignment} & Yes & \fullmark & \partmark & \partmark & \partmark & \fullmark & \fullmark \\
    RSIGym~\citep{meng2026rsigym} & Yes & \fullmark & \nomark & \partmark & \fullmark & \partmark & \partmark \\
    \midrule
    AIRA$_2$~\citep{hambardzumyan2026aira2} & \nomark & \nomark & \nomark & \nomark & \fullmark & \fullmark & \partmark \\
    CORAL~\citep{qu2026coral} & Optional & \nomark & \nomark & \nomark & \fullmark & \partmark & \partmark \\
    Arbor~\citep{jin2026arbor} & \nomark & \partmark & \nomark & \nomark & \partmark & \partmark & \fullmark \\
    Praxist~\citep{li2026praxist} & \nomark & \partmark & \nomark & \nomark & \partmark & \partmark & \fullmark \\
    \keyrow
    {\bfseries Athanor} & {\bfseries Read only} & \fullmark & \fullmark & \fullmark & \fullmark & \fullmark & \fullmark \\
    \bottomrule
  \end{tabularx}
  \par\smallskip
  \begin{minipage}{\textwidth}
  \small
  \fullmark\ explicit support; \partmark\ partial or task-dependent support;
  \textcolor{seedgray}{--} not supported or not reported.
  A generalist checkpoint targets multiple capabilities in one model.
  Checkpoint lineage records parent--child relationships across training stages;
  research lineage links hypotheses, artifacts, and evidence. Sealed final evaluation means that the final evaluation is kept apart from the feedback the search receives. A system gets partial support if it isolates the evaluator, or if it uses held-out checks while it selects among candidates, but does not seal a final evaluation.

  $^{*}$A-Evolve-Training reports public-leaderboard retrieval; broader access
  is unspecified. CORAL disables Internet in its main comparisons and reports
  web-enabled variants separately. RSIGym runs its final evaluation under a
  fixed protocol on tasks the agents have seen during development, and reports
  one variant without Internet access.
  \end{minipage}
\end{table}

{\small
\setlength{\tabcolsep}{4pt}
\begin{longtable}{@{}
  >{\raggedright\arraybackslash}p{0.28\textwidth}
  >{\raggedright\arraybackslash}p{0.22\textwidth}
  >{\raggedright\arraybackslash}p{0.26\textwidth}
  >{\raggedright\arraybackslash}p{0.19\textwidth}@{}}
  \caption{\textbf{Automated post-training settings.} Agent control ranges from a
  single specialized training intervention to sustained recipe search.}
  \label{tab:automated-posttraining}\\
  \toprule
  \tablehead
  Work or family & Evolved or graded artifact & Search and evaluation boundary & Persistent evidence \\
  \midrule
  \endfirsthead
  \multicolumn{4}{l}{\footnotesize\emph{Automated post-training settings (continued)}}\\
  \toprule
  \tablehead
  Work or family & Evolved or graded artifact & Search and evaluation boundary & Persistent evidence \\
  \midrule
  \endhead
  \bottomrule
  \endfoot
  PostTrainBench~\citep{ptb} & Data, code, and recipe for one specialist model & Internet access; one benchmark, one GPU, ten hours & Submitted workspace and artifacts \\
  \midrule
  Locus~\citep{locus} & Specialist post-training recipe & Internet access; parallel, multi-day PostTrainBench search & Experiment records (agent traces, training jobs, model artifacts) \\
  \midrule
  AutoTrainess~\citep{yu2026autotrainess} & PostTrainBench specialist recipe & Workflow interfaces for planning, data, training, and evaluation & Explicit plans and experiment logs \\
  \midrule
  Specialized training benchmarks~\citep{ftdojo,agent2rlbench,rsibenchdata} & Domain fine-tuning, RL pipelines, or data interventions & Task-specific training and feedback under controlled budgets & Submitted models, code, data, and trajectories \\
  \midrule
  Method-specific search systems~\citep{du2026datamaster,fang2026llmzero,chen2026evotrainer} & Training data, RL strategies, or policy--harness pairs & Search focused on a selected part of the post-training pipeline & Data or strategy trees, memory, and evolution histories \\
  \midrule
  A-Evolve-Training~\citep{shi2026evolve} & One 30B model (also run at 120B and 550B) and its data/recipe & Four autonomous rounds with development and challenge scores & Auditable multi-round trajectory \\
  \midrule
  AAR~\citep{chen2026automatedalignment} & Post-training method combinations, data, and model weights & Internet access; multi-day parallel safety search with capability checks and separate generalization tests & Pre-result method reports, code snapshots, shared findings, and scores \\
  \midrule
  RSIGym~\citep{meng2026rsigym} & A model fine-tuned with LoRA and the harness that runs it, one system per benchmark & Remote training, inference, evaluation, and sandbox services under a per-run budget; frontier models may generate training data; final scores are on tasks seen during development & Service-side records of training runs and evaluation jobs, a cost ledger, and trajectories \\
  \midrule
  \keyrow
  {\bfseries Athanor} & One generalist checkpoint and its recipe lineage, built by a team of agents & Read-only Internet access; multi-benchmark feedback; held-out twin suite; days on eight GPUs per agent & Journal, checkpoint lineage, transcripts, and notes \\
\end{longtable}
}

\section{Experimental Details}\label{app:campaign-protocol}
This appendix gives the full setting behind \cref{sec:setting}.

\paragraph{Setting.}
\Cref{tab:c1-setting} lists what each agent received and what it was not
allowed to do, and \cref{tab:suites} lists the two suites. The campaign protocol fixes
decoding: mathematics averages four samples per problem (temperature 0.6,
top-$p$ 0.95), and the other tasks decode greedily. Mathematics, logic, and code
get 16,384 output tokens, knowledge and instruction following get 8,192, and
ALFWorld gets 256 per step. Under this protocol, every reference model was measured with two seeds. The default starter was run and measured three times.

\begin{table}[ht]
  \centering
  \small
  \setlength{\tabcolsep}{5pt}
  \caption{\textbf{Setting of the campaign.} Every setting is recorded with each result; settings changed during the campaign give the hour (\cref{tab:timeline}).}
  \label{tab:c1-setting}
  \begin{tabularx}{\textwidth}{@{}>{\raggedright\arraybackslash}p{0.2\textwidth}>{\raggedright\arraybackslash}X@{}}
    \toprule
    \tablehead
    Aspect & Setting \\
    \midrule
    Researchers & Four agents, Claude Opus 5.5 at extra-high (xhigh) reasoning effort in Claude Code; no skills (packaged instructions), research advice, or findings from other campaigns \\
    \midrule
    Base and main reference & Qwen3-4B-Base; Qwen3-4B-Instruct-2507, the vendor's July 2025 update of Qwen3-4B \\
    \midrule
    Resources & One node per agent, a GPU server with eight GPUs (80\,GB of HBM3 memory each) and a high-bandwidth GPU interconnect; a research window planned for 72 hours, extended to 144 hours during the campaign, and closed at 104 \\
    \midrule
    Team & The agents read each other's workspaces, results, and journals and fetch each other's checkpoints; the result counts the team's submission, which the agents are told at hour 15.8 \\
    \midrule
    Data & The decontaminated pool of 447 public datasets; further public datasets, fetched through the environment and decontaminated on entry; problems the agents generate; and samples from checkpoints they have trained \\
    \midrule
    Network & The agents can read web pages but cannot post, and they can reach hosts for software packages, papers, and code. Claude Code's built-in tools for web search, waiting, and notifications are turned on at hour 73 \\
    \midrule
    Not allowed & Calling or downloading any other model as a teacher, the vendor's weights included; writing answers or labels by hand \\
    \midrule
    Visible references & The agents see the development scores of the base, the vendor's model, and the default starter. For the vendor's model they also see its score on each item, but not its answers \\
    \midrule
    Humans & Humans give no research advice. The operator replaces nodes that the platform takes back, and sends messages that state facts about the setup or about the agents' own records \\
    \bottomrule
  \end{tabularx}
\end{table}

\begin{table}[ht]
  \centering
  \small
  \setlength{\tabcolsep}{5pt}
  \caption{\textbf{The two suites.} The agents evaluate on the development
  suite and see every scored answer; the held-out suite is unknown to them and
  is scored after the campaign. Counts are items (ALFWorld: games).}
  \label{tab:suites}
  \begin{tabularx}{\textwidth}{@{}>{\raggedright\arraybackslash}p{0.21\textwidth}>{\raggedright\arraybackslash}X>{\raggedright\arraybackslash}X@{}}
    \toprule
    \tablehead
    Capability & Development (visible) & Held-out (sealed) \\
    \midrule
    Mathematics & Five 2025 competitions including AIME and HMMT February (183) & AIME 2026, HMMT February 2026, HMMT November 2025 (93) \\
    Logic & ZebraLogic, one half (500) & ZebraLogic, the other half (500) \\
    Code & LiveCodeBench, problems from 2024 (511) & LiveCodeBench, problems from 2025 (175) \\
    Knowledge & MMLU-Pro (2,000) & SuperGPQA (1,000) \\
        Instruction following & IFEval (541) & IFBench (300) \\
    Multi-step action & ALFWorld, four task types (91) & ALFWorld, two unseen task types (85) \\
    \bottomrule
  \end{tabularx}
\end{table}

\paragraph{Data pool.}
The pool holds 447 public datasets (5,129 files, about 384 million rows), of
which the researchers see the name, size, format, and source. Before the
campaign, every row was checked against the items of both suites and removed on
overlap (\cref{app:system-details}); 40,113 rows were removed. The agents
passed their own datasets through the same check 12 times, for generated logic
puzzles and ALFWorld trajectories.
\Cref{fig:data-pool-coverage} shows how much of the pool the team used. The fine-tuning data draw on
AM-Thinking-v1-Distilled, Mixture-of-Thoughts, the Dolci and Nemotron
post-training sets, OpenCodeReasoning-2, and ALFWorld trajectories; the
reinforcement-learning problems draw on Skywork-OR1, Polaris, AceReason-Math, the
Nemotron reinforcement-learning sets for multiple-choice questions and
instruction following, \mbox{IF\_multi\_constraints\_upto5}, DeepCoder, and a pool of logic
grid puzzles.

\begin{figure}[ht]
  \centering
  \includegraphics[width=0.8\textwidth]{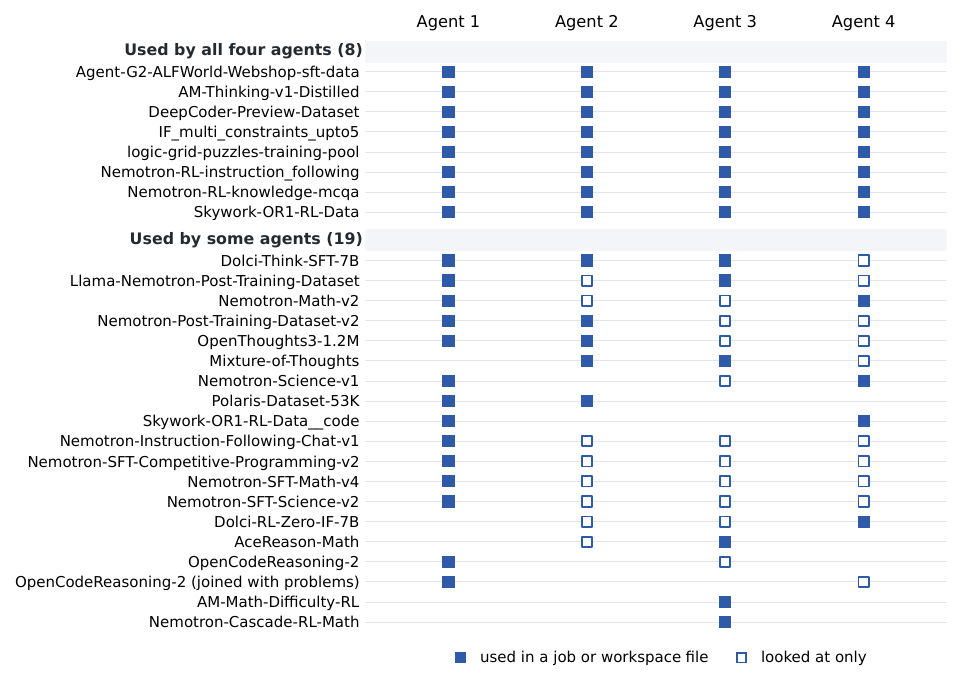}
  \caption{\textbf{Datasets the agents worked with.} Of the 447 datasets in
  the pool, the agents looked at 43 (named in a shell command, a workspace
  file, or a job) and used the 27 shown here in a job or a workspace file. A
  filled square is a dataset the agent used, and an open square is one it only
  looked at. Eight datasets are used by all four agents.
  OpenCodeReasoning-2 appears twice because the pool holds the original and a
  copy joined with its problem statements.}
  \label{fig:data-pool-coverage}
\end{figure}

\begin{table}[ht]
  \centering
  \small
  \caption{\textbf{Timeline of the campaign.} Hours from the start of the
  research window.}
  \label{tab:timeline}
  \begin{tabularx}{\textwidth}{@{}l X@{}}
    \toprule
    \tablehead
    Hour & Event \\
    \midrule
    0 & The four agents start, each on its own node \\
    \midrule
    15.8 & The operator sends a notice that the final result counts the team's submission. The protocol had fixed this before the start, but the agents' documents did not mention it \\
    \midrule
    46, 70, 94 & The platform reclaims a node; the agent's session resumes on a new node from its record, with the outage added to its window \\
    \midrule
    59 & The agents are told that the research window is extended from 72 to 144 hours \\
    \midrule
    73 & Claude Code's built-in tools for web search, waiting, and notifications are turned on \\
    \midrule
    97 & An operator message gives each agent statistics from its own evaluation records: how many of its answers end at the output limit (\cref{app:plateau}); all four final submissions were made before it \\
    \midrule
    104 & The campaign closes before the end of the extended window, because the best development score has risen by less than one point in the previous day \\
    \bottomrule
  \end{tabularx}
\end{table}

\paragraph{Selection and main result.}
Each agent submits one checkpoint before the window closes. The team's
submission is the submitted checkpoint with the highest full development-suite
score; the held-out suite plays no part in the choice. The main result is that
submission's score on the held-out suite, the plain mean of six tasks: 46.7 under the campaign protocol (two seeds) and 48.2 in the final measurement (four seeds, \cref{tab:v4-tasks}). \Cref{tab:sealed} gives the
held-out scores of all four final submissions. The rule selects \agent{3}'s submission. \agent{2}'s submission scores 0.01 development points lower. Under the campaign protocol, \agent{2}'s submission scores 0.5 held-out points higher (standard error of the difference 0.7), so the held-out suite does not separate them.

\begin{table}[tb]
  \centering
  \small
  \setlength{\tabcolsep}{5pt}
  \caption{\textbf{The four final submissions, under the campaign protocol.} Held-out columns as in \cref{tab:c1-main}. The development mean is shown with the number of seeds in parentheses. The held-out scores come from two seeds each, run after the campaign. The standard error of each held-out mean is 1.3 to 1.4.}
  \label{tab:sealed}
  \begin{tabular}{@{}lc|ccccccc@{}}
    \toprule
    \tablehead
    & Development & \multicolumn{7}{c}{Held-out suite} \\
    \tablehead
    Submission & \textbf{Mean} & Mat & Log & Cod & Kno & Ins & Act & \textbf{Mean} \\
    \midrule
        \keyrow \agent{3} (the team's submission) & 66.1 (3) & 29.7 & 63.7 & 34.9 & 36.0 & 37.0 & 78.8 & 46.7 \\
    \agent{2} & 66.1 (3) & 30.1 & 63.4 & 36.3 & 37.2 & 36.7 & 79.4 & 47.2 \\
    \agent{1} & 66.0 (4) & 29.6 & 61.8 & 35.2 & 37.1 & 39.9 & 79.4 & 47.1 \\
    \agent{4} & 65.5 (3) & 30.5 & 61.4 & 34.0 & 36.4 & 38.0 & 75.3 & 45.9 \\
    Instruct-2507 & 66.4 (2) & 42.2 & 81.0 & 31.4 & 36.4 & 31.1 & 65.9 & 48.0 \\
    \bottomrule
  \end{tabular}
\end{table}

\begin{table}[tb]
  \centering
  \small
  \setlength{\tabcolsep}{3.6pt}
  \caption{\textbf{The four final submissions in the final measurement.} Columns as in \cref{tab:c1-main}, whose bottom half gives the other references. Four seeds, up to 32,768
  output tokens, temperature 0.6. Standard errors of the means are 0.5--0.9 on
  the development suite and 0.6--1.3 on the held-out suite.}
  \label{tab:v4-tasks}
  \begin{tabular}{@{}lccccccc|ccccccc@{}}
    \toprule
    \tablehead
    & \multicolumn{7}{c|}{Development suite} & \multicolumn{7}{c}{Held-out suite} \\
    \tablehead
    Submission & Mat & Log & Cod & Kno & Ins & Act & \textbf{Mean} & Mat & Log & Cod & Kno & Ins & Act & \textbf{Mean} \\
    \midrule
        \keyrow \agent{3} (the team's submission) & 38.4 & 63.0 & 51.1 & 70.4 & 85.2 & 95.6 & 67.3 & 33.3 & 62.5 & 37.9 & 43.3 & 37.6 & 74.7 & 48.2 \\
        \agent{2} & 38.1 & 63.8 & 49.8 & 70.1 & 84.0 & 96.7 & 67.1 & 31.2 & 62.8 & 37.4 & 42.8 & 36.8 & 70.9 & 47.0 \\
    \agent{1} & 38.8 & 63.6 & 50.4 & 69.9 & 84.0 & 96.4 & 67.2 & 29.8 & 62.5 & 37.3 & 42.6 & 38.1 & 73.5 & 47.3 \\
    \agent{4} & 37.3 & 60.5 & 50.5 & 69.8 & 85.0 & 93.1 & 66.0 & 32.8 & 60.9 & 37.7 & 42.0 & 38.8 & 65.9 & 46.3 \\
    Instruct-2507 & 44.2 & 79.5 & 45.1 & 64.1 & 84.1 & 81.8 & 66.5 & 42.5 & 77.7 & 33.6 & 36.5 & 30.3 & 66.8 & 47.9 \\
    \bottomrule
  \end{tabular}
\end{table}

\paragraph{Campaign timeline.}
\Cref{tab:timeline} lists what happened during the campaign.

\paragraph{Contamination audit.}
After the campaign, every text that entered the four agents'
sessions during the campaign was scanned: none
matches a held-out item, and no agent visited a site that hosts a held-out
dataset. The ALFWorld games the agents can play contain only the four development task types, so no trajectory they generate or replay is of a held-out type. The public ALFWorld fine-tuning data they use holds all six types, but the agents read it only through the starter's generator, which replays each trajectory in its game and so drops the 1,346 trajectories of the two held-out types, whose games are absent. We also scanned the datasets the agents built themselves: of the 108 datasets the
team's submission descends from, the 61 that could be scanned in full contain
no held-out item; 30 others were lost with the nodes' local disks before the final scan, and 17 exceeded the scan's time or row limit; these 47 are reformatted copies and mixtures of datasets from the decontaminated pool, logic puzzles, and samples from the team's checkpoints. The fine-tuning mixtures of \agent{1} and \agent{3},
rebuilt from the recipe, contain no held-out item either.

\paragraph{The final measurement.}
Under the campaign protocol the two reasoning releases score below Instruct-2507: ALFWorld allows 256 tokens per step. Thinking-2507 starts each step with reasoning, and the 256 tokens run out before it chooses an action, so it scores zero on ALFWorld; the hybrid Qwen3-4B, which also reasons first, scores 24.2. Two thirds of Thinking-2507's mathematics answers are also cut off at 16,384 tokens. The final measurement, added after the campaign for this reason, samples every model with the setting Qwen recommends for Qwen3 in thinking mode
(temperature 0.6, top-$p$ 0.95, top-$k$ 20, one sample per item). It allows 32,768 output tokens on mathematics, logic, code, and knowledge, and 8,192 on instruction following and per ALFWorld step. Every model is prompted through one shared Qwen3 chat template with its default settings, so the hybrid Qwen3-4B writes out its reasoning under both protocols. \Cref{tab:v4-tasks} gives the four final submissions under this protocol. Each model was evaluated with four seeds and with one and the same build of the evaluator. To check that this setting does not disadvantage Instruct-2507, we also sample it with its own recommended settings (temperature 0.7, top-$p$ 0.8, top-$k$ 20) and its own chat template, on the same items and output limits but with a later build of the evaluator. With its own sampling it scores 66.4 and 47.5, no higher than in \cref{tab:v4-tasks}. In the final measurement (\cref{tab:v4-tasks}), all four final submissions
score within 1.6 points of Instruct-2507 on both suites and trail the reasoning releases (\cref{tab:c1-main}, bottom).

\FloatBarrier
\section{Additional Results}\label{app:additional-results}
This appendix gives the evidence behind the main text. \cref{app:plateau} supports the task profile and the climb of \cref{sec:results,sec:f-slow}; \cref{app:lineage} supports \cref{sec:f-recombination,sec:f-convergence,sec:f-data}; \cref{app:team} supports \cref{sec:f-team}; \cref{app:data-sense} supports \cref{sec:f-data}; \cref{app:lanes} supports \cref{sec:execution}; and \cref{app:posttrain-science} supports \cref{sec:insights}.

\subsection{Why the Submission Trails on Logic and Mathematics}\label{app:plateau}

On logic and mathematics the team's models score below Instruct-2507
(\cref{sec:results}; \cref{fig:c1-tasks} shows how each task climbs). On logic the cause is the large puzzles, which the models rarely solve; under the campaign protocol this shows as answers cut off at the output limit. We measured this on the development suite, from the
record of every answer in two evaluation seeds of the team's submission and of Instruct-2507.

\begin{figure}[H]
  \centering
  \includegraphics[width=\textwidth]{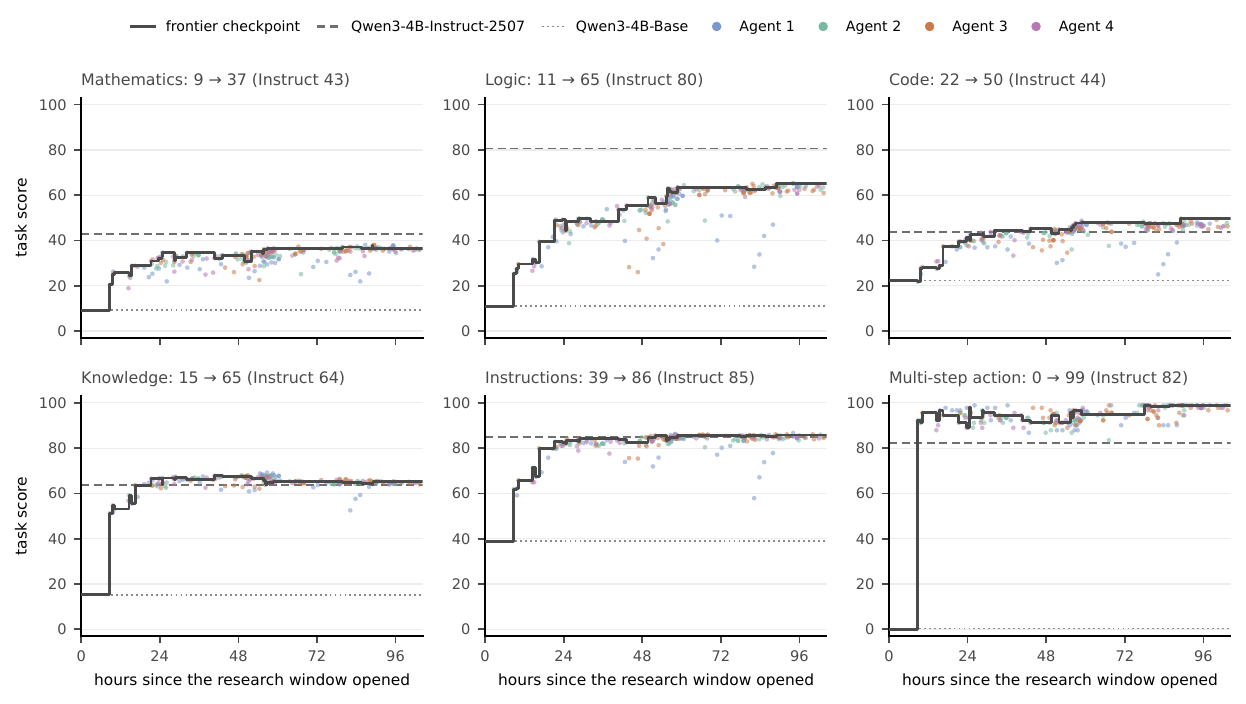}
  \caption{\textbf{How each development task climbs.} Multi-step action passes Instruct-2507 by hour 9 and knowledge by hour 21; instruction following comes within a point of it by hour 29, and code passes it by hour 32. After the first day, the largest gain is on logic, from 49 to 65, still below Instruct-2507's 80; mathematics stays below throughout. Each panel shows every evaluation's score on one task (dots, colored by agent) and, as a staircase, the task score of the checkpoint that holds the best development mean so far (the frontier of \cref{fig:c1-climb}). Dashed and dotted lines are Instruct-2507 and the base under the campaign protocol. The staircase uses single evaluations of the leading checkpoint, so its endpoints are not the submission's multi-seed means of \cref{tab:c1-main}.}
  \label{fig:c1-tasks}
\end{figure}

On logic, 17.5\% of the answers of the team's submission use up the whole
16,384-token output limit before reaching an answer, and
each of these answers scores zero. Instruct-2507 has no separate thinking phase and writes
shorter answers: only 0.8\% are cut off this way. On
mathematics the corresponding shares are 24.2\% for the team's submission and 9.8\%
for Instruct-2507. Among the answers that do finish, the team's submission scores 76.4\% on logic and 50.1\% on mathematics, against 81.0\% and 47.2\% for Instruct-2507; on logic its finished answers are mostly the smaller puzzles. Cut-off answers concentrate in the large puzzles (\cref{fig:c1-plateau}).

The agents noticed this problem themselves, before the operator's message at hour 97 that gave each agent its own share of answers cut off (\cref{tab:timeline}). This is why their
self-distillation keeps only the shortest correct answer to each problem. Four
rounds of fine-tuning on the shortest correct samples (\cref{tab:lineage-steps}) left the share of cut-off logic answers of the four final submissions at 17.5--18.9\%, against 17.4\% (87 of 500) for \agent{3}'s extrapolation before the first round (\cref{tr:rft}). The
reason is that the model is rarely right on the large puzzles, so on those
puzzles there are few correct answers to learn from (\cref{fig:c1-plateau}), and few short ones among them (\cref{tr:compact}). The training pools lean toward small puzzles: \agent{3} keeps only puzzles of at most 16 cells for reinforcement learning (\cref{tr:turn}).

Raising the output limit does not recover these answers. In the final
measurement, with an output limit of 32,768 tokens and sampled decoding, the team's submission's logic score changes by less than one point (63.4 to 63.0), and the four final submissions remain 15--19 points below Instruct-2507 on logic
(\cref{tab:v4-tasks}). On mathematics, the answers that finish are at least as accurate as Instruct-2507's, so under the campaign protocol much of the gap shows up as truncation; that the gap persists with the longer limit (38.4 against 44.2) shows that truncation is not all of it, and we have not identified the rest.

\begin{figure}[H]
  \centering
  \includegraphics[width=0.6\textwidth]{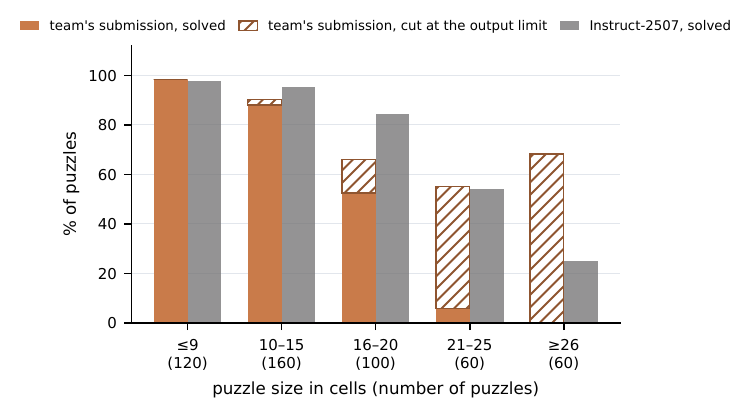}
  \caption{\textbf{Logic puzzles by size} (campaign protocol). Share of the 500 development logic puzzles that the team's submission and Instruct-2507 solve, grouped by puzzle size. Size is the number of cells (houses $\times$ attributes). Each team bar stacks the share solved (solid) and the share cut off at the output limit before an answer (hatched).}
  \label{fig:c1-plateau}
\end{figure}

\subsection{The Winning Lineage}\label{app:lineage}

\Cref{tab:lineage-steps} lists the training and merge steps from the base to the team's submission. Some rows combine several rounds of the same stage.  Each score is a development mean, and the number of seeds is given in parentheses.

\begin{table}[t]
  \centering
  \small
  \setlength{\tabcolsep}{4pt}
  \caption{\textbf{Steps of the winning lineage.} $R_i$: Agent $i$'s
  reinforcement-learning model ($R_4'$: a later checkpoint of the same run);
  base: Qwen3-4B-Base.}
  \label{tab:lineage-steps}
  \begin{tabularx}{\textwidth}{@{}>{\raggedright\arraybackslash}p{0.22\textwidth}>{\raggedright\arraybackslash}X>{\raggedright\arraybackslash}p{0.15\textwidth}>{\raggedright\arraybackslash}p{0.16\textwidth}@{}}
    \toprule
    \tablehead
    Step (agent) & Operation & GPU h & Dev mean (seeds) \\
    \midrule
    SFT (\agent{1}, \agent{3}, \agent{4}) & 369k, 602k, 501k rows; lr $5\times10^{-5}$, sequences up to 16,384 tokens, one epoch & 51.1, 67.7, 66.2 & 46.8, 49.7, 48.0 (1) \\
    GRPO, $R_1$ (\agent{1}) & Five rounds; greedy share 0.25 in the second and third, 0.15 after; rollouts to 16,384 tokens; more logic puzzles in the last & 151 & 59.6 (1) \\
    GRPO, $R_3$ (\agent{3}) & Four rounds; greedy share 0.25 from the second & 86.5 & 59.6 (1) \\
    GRPO, $R_4$ (\agent{4}) & Average of its own two fine-tuned models (the second is not listed above), then one long run with greedy share 0.25 & 68.5 & 61.7 (1) \\
    Average (\agent{1}) & $0.5R_1+0.5R_3$ & 0 & 61.8 (1) \\
    Weighted average (\agent{4}) & $0.25R_1+0.25R_3+0.5R_4$ (the merged model of \cref{fig:c1-science}b) & 0 & 63.3 (1) \\
    Extrapolation (\agent{3}) & $0.3R_1+0.3R_3+0.6R_4-0.2\,\text{base}$, that average at $f=1.2$ & 0 & 64.0 (4) \\
    Extrapolation (\agent{4}) & $0.65R_4'+0.65(0.5R_1+0.5R_3)-0.3\,\text{base}$ ($f=1.3$) & 0 & 65.5 (3) \\
    Self-distillation (\agent{3}) & Two rounds on \agent{3}'s extrapolation; shortest correct of four samples & 86.3 & 65.0, 64.8 (3) \\
        Fine-tuning on samples (\agent{2}) & One round on \agent{4}'s extrapolation, with the samples \agent{3} drew from its own extrapolation & 35.0 & 65.3 (3) \\
    Average (\agent{2}) & \agent{2}'s fine-tuned model and \agent{3}'s self-distilled model, half each & 0 & 65.9 (3) \\
        Self-distillation (\agent{2}) & A further round on that average & 27.8 & 66.1 (3) \\
    Average (\agent{3}) & \agent{2}'s second round and the average of \agent{3}'s two rounds (65.2), half each: the team's submission & 0 & 66.1 (3) \\
    \bottomrule
  \end{tabularx}
\end{table}

\paragraph{Data.}
The fine-tuning mixtures of Agents 1, 3, and 4, which lead to the submission, differ in their supplementary data: verified competitive code, science
questions rewritten as multiple choice, instruction-following data with
checkable constraints, chat, and ALFWorld expert trajectories repeated two or
three times. Reinforcement learning used 20--27 thousand problems per round from
public mathematics, code, instruction, and multiple-choice sets and from
generated logic grids, with the logic share raised over the rounds. All rewards
are rule-based: 1 for a correct answer, 0 otherwise.

\paragraph{What was tried and dropped (findings (3) and (4)).}
The agents tried and dropped several things, and each of the following is one
attempt. Fine-tuning a model on about 6,900 solver traces for logic puzzles
lowered its score on a logic-only diagnostic evaluation from about 49 to 27 (\cref{tr:solver}). A
reinforcement-learning run with many logic puzzles lowered the validation
accuracy that the training job itself measures, for logic from 0.67 to 0.61 and
for mathematics from 0.52 to 0.45, within ten steps, and the agent cancelled
it. A second fine-tuning on shorter answers lowered the mathematics score from
25.0 to 18.9. In a reinforcement-learning run with a penalty on answer length,
the development score went from 65.1 at step 10 to 64.6 at step 20.  TIES merging~\citep{yadav2023ties}, tried twice on the parents of the extrapolations, scored
64.5 and 64.7, below the best linear extrapolation (65.5).

\Needspace{6\baselineskip}
\subsection{Exchanges Between Agents}\label{app:team}

\Cref{tab:team-events} lists a selection of the transfers of ideas and checkpoints between agents
that the record confirms.

\begin{table}[H]
  \centering
  \small
  \setlength{\tabcolsep}{4pt}
  \caption{\textbf{A selection of ideas and checkpoints that moved between agents.} Hours from the start of the research window, from when the source agent has the idea or checkpoint to when a teammate takes it up (a single time: the fetch is the take-up; a semicolon separates two transfers in one row); each row is confirmed by a fetch, a job's inputs, or a
  note in the receiving agent's workspace.}
  \label{tab:team-events}
  \begin{tabularx}{\textwidth}{@{}rp{0.19\textwidth}p{0.14\textwidth}X@{}}
    \toprule
    \tablehead
    \# & Hours & From $\rightarrow$ to & What moved \\
    \midrule
    1 & 12.9 $\rightarrow$ 16--18.4 & 3 $\rightarrow$ 1, 4, 2 & Greedy rollouts in reinforcement learning; \agent{4}'s notes credit \agent{3} \\
    \midrule
        2 & 14.1, 15.5 $\rightarrow$ 15.8 & 1, 2 $\rightarrow$ 4 & Two checkpoints, used by \agent{4} in its merge of teammates' models at hour 15.9 (not on the submission's lineage) \\
    \midrule
    3 & 18.7 & 3 $\rightarrow$ 2 & A reinforcement-learning checkpoint, used as a new starting point \\
    \midrule
    4 & 22.3 $\rightarrow$ 37.2 & 3 $\rightarrow$ 1 & $R_3$, averaged with \agent{1}'s own model \\
    \midrule
    5 & 42.8; 51.2 & 1 $\rightarrow$ 4; 4 $\rightarrow$ 3 & \agent{1}'s average and \agent{4}'s $R_4$, which \agent{4} merges into the weighted average that \agent{3} extrapolates \\
    \midrule
        6 & 55.8 $\rightarrow$ 57.0 & 3 $\rightarrow$ 4 & The extrapolated merge; \agent{4}'s job note names \agent{3} \\
    \midrule
    7 & 59 $\rightarrow$ 67.3; 59 $\rightarrow$ 82.3 & 3 $\rightarrow$ 2; 3 $\rightarrow$ 4 & The self-distillation pipeline; \agent{2}'s script names its source \\
    \midrule
    8 & 60.8 & 4 $\rightarrow$ 2 & \agent{4}'s extrapolated model, which \agent{2} fine-tunes on samples \\
    \midrule
    9 & 76.4 & 3 $\rightarrow$ 2 & \agent{3}'s second self-distilled model, averaged a minute later \\
    \midrule
        10 & 83.9 $\rightarrow$ 86.5--87.9 & 2 $\rightarrow$ 4, 3, 1 & \agent{2}'s second self-distilled model, fetched by all three \\
    \bottomrule
  \end{tabularx}
\end{table}

\subsection{How the Agents Inspected Data}\label{app:data-sense}
We counted the commands in each agent's transcripts that displayed rows of data. These counts are lower bounds: we checked each command that we counted, but other commands may also have displayed rows. In the first six hours each agent displayed about 30--70 row excerpts. In the rest of the campaign each displayed 2--12. Each agent also ran at least six checks on the content of a dataset or on how the scorer grades the rows, and several checks that only computed statistics, such as counts and lengths. Besides the catches of \cref{sec:f-data}, they found answer formats the evaluator would not parse. Correctness was checked less:
one agent left a suspicious ground truth to a filter it did not run, and a
converter marked solutions verified without comparing them to the expected
answer.

\subsection{The Campaign Agent by Agent}\label{app:lanes}
\Cref{fig:c1-lanes} shows the campaign of \cref{fig:c1-execution} separately for each agent. Each agent's GPUs stay busy through the phases of the recipe: fine-tuning in the first hours, then GRPO, and later mostly self-distillation sampling for Agents 2 to 4, with evaluations in between. The agents' turns come in bursts as results arrive. The figure's top panel shows single evaluations, which run high (\cref{fig:c1-science}a); the multi-seed means are in \cref{tab:sealed}.

\begin{figure}[tb]
  \centering
  \includegraphics[width=\textwidth]{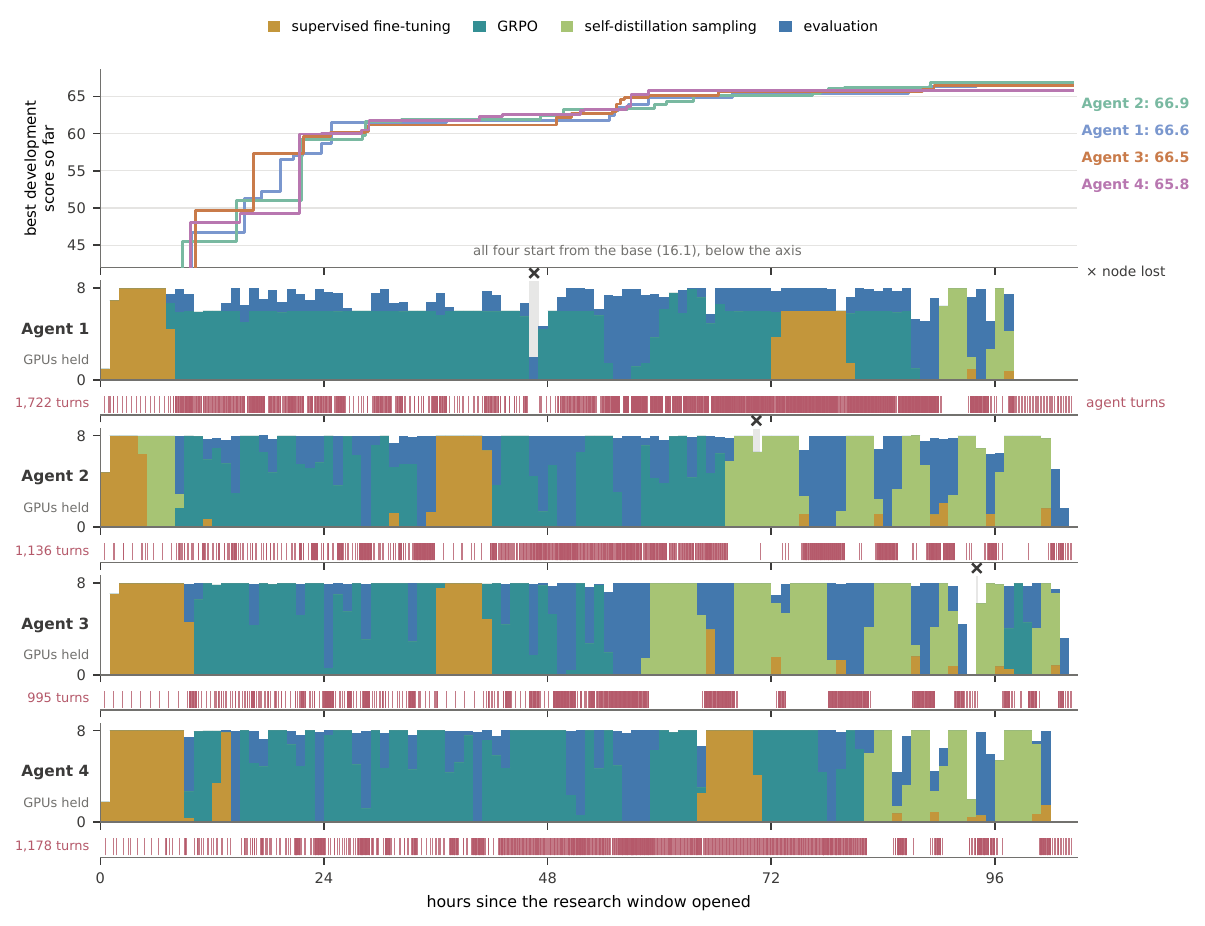}
  \caption{\textbf{The campaign agent by agent.} Top: each agent's best single evaluation so far; the lead changes hands several times, and all four end between 65.8 and 66.9. Below, one lane per agent: the GPUs it holds, as hourly means by phase of the recipe, and one tick per agent turn. A shaded span and a cross mark a node the platform took back.}
  \label{fig:c1-lanes}

\end{figure}

\subsection{Post-Training Observations}\label{app:posttrain-science}
Post-training itself remains an open scientific problem: data composition
interacts with initialization, objective, stage order, and checkpoint
composition. The record of the campaign holds such interactions as measured
lineages. Each of the following is a single observation.
When two reinforcement-learning models of
equal strength (near 57) from different lineages are averaged, the average
scores about 2.7 points higher than the better parent (\cref{tr:soup}). When a strong model is
averaged with weaker ones, the average scores no higher than the strong model
alone (an observation in \agent{4}'s notes; \cref{tr:soup}). When \agent{3} moves a merged model further from the base, its logic
score rises as the factor grows from 1.0 to 1.2 (56.4, 59.6, 63.0 at factors
1.0, 1.1, 1.2), and its knowledge score falls once the factor passes 1.1 (66.8,
65.7, 63.8 at factors 1.1, 1.2, 1.3); the best factor therefore depends on
which task matters more (\cref{fig:c1-science}b). Self-distillation (training a model on its own
sampled answers) raises its score by 1.0 point (three seeds) in its first round; later rounds change it by at most 0.2 (\cref{tab:lineage-steps}). Fine-tuning a different merge on the same samples does not help on its own:
the result scores 65.3, against 65.5 for the merge before, although averaging it with \agent{3}'s self-distilled model gives 65.9 (\cref{tab:lineage-steps}).

\FloatBarrier
\section{Research Trajectories}\label{app:trajectories}

This appendix shows the agents' own work behind the findings of
\cref{sec:discoveries}. Each quote is copied from an agent's notes file or
from the file it writes for its teammates; r1--r4 in the quotes are Agents 1--4, and an ellipsis marks an omission. In the
agents' shorthand, dev6 is the development suite; zebra, mmlu (or mcq), ifeval, and
alfworld are its logic, knowledge, instruction-following, and multi-step
action tasks; a soup is an average of checkpoints; diag is a diagnostic evaluation on some of the tasks; RFT (also expert iteration) is
rejection-sampling fine-tuning, the self-distillation of \cref{sec:f-recombination}; RESP is the length limit of one answer in a training rollout; names such as s3Lm or soup-e are the agents' checkpoint names; official is Instruct-2507; and the other abbreviations (T, sN, val) are as in \cref{sec:f-recombination}.

\subsection{A Complete Agent Turn}\label{app:episode}
The research episode shown in \cref{sec:f-recombination} turns on one step, in
which \agent{3} replaces sampled rollouts with a mix of sampled and greedy ones.
The box below shows that step as it appears in the agent's transcript. It lists the shell commands that the agent issued in a few minutes at hour 12. The list starts when a validation result arrives and ends when the agent stops its training job. Commands
and outputs are copied from the transcript; an ellipsis marks an omission, and
the file-reading commands of step 2 are summarized.

\begin{agenttrace}[label={tr:turn}]{One agent turn, from a validation result to a new rollout loop (\agent{3}, hour 12)}
\small
\textbf{Woken by.} Earlier, the agent had set up a watch on its training log, a program that wakes the agent when a given event appears in the log. The watch wakes the agent now, because the checkpoint at step 25 of its first GRPO job has been validated using greedy decoding.

\noindent\begin{minipage}{\linewidth}
\begin{lstlisting}[style=agentnote]
val-core/ifeval/acc/mean@1:0.296875        val-aux/ifeval/truncated/mean@1:0.40625
val-core/zebralogic/acc/mean@1:0.09375     val-aux/zebralogic/truncated/mean@1:0.859375
val-core/math_2025/acc/mean@1:0.234375     val-aux/math_2025/truncated/mean@1:0.625
val-core/code/acc/mean@1:0.21875           val-aux/code/truncated/mean@1:0.6875
val-core/mmlu_pro/acc/mean@1:0.421875      val-aux/mmlu_pro/truncated/mean@1:0.3125
\end{lstlisting}
\end{minipage}

\textbf{1. Classify the cut answers.} The agent writes a short program. It compares the validation answers before training (step 0) with those at step 25. For each task it counts how many answers were cut off at the output limit, and how many of those cut-off answers end in a repetition loop. The program calls an answer a loop if its last 6,000 characters shrink to under 15\% of their size when compressed.

\begin{lstlisting}[style=agentnote]
$ python3 - <<'EOF'
for step in (0, 25):
    agg = collections.defaultdict(collections.Counter)
    for line in open(f'/scratch/grpo/grpo1/val/{step}.jsonl'):
        r = json.loads(line); t = task_of(r); a = agg[t]
        a['n'] += 1; a['acc'] += r['acc']
        if r['truncated'] or r.get('over_budget'):
            a['trunc'] += 1
            tail = r['output'][-6000:].encode()
            a['loop'] += len(zlib.compress(tail)) / max(1, len(tail)) < 0.15
    print(step, {t: (a['n'], round(a['acc']/a['n'],2), a['trunc'], a['loop']) for t, a in sorted(agg.items())})
EOF
0 {'code': (32, 0.19, 24, 19), 'ifeval': (64, 0.31, 24, 22), 'math': (64, 0.36, 35, 12), 'mcq': (64, 0.41, 20, 6), 'zebra': (64, 0.08, 51, 33)}
25 {'code': (32, 0.22, 22, 17), 'ifeval': (64, 0.3, 26, 26), 'math': (64, 0.23, 40, 9), 'mcq': (64, 0.42, 21, 8), 'zebra': (64, 0.09, 55, 33)}
\end{lstlisting}

Each tuple is (answers, accuracy, cut at the limit, loops among the cut); the program also counts answers over the token limit as cut.
After 25 steps of GRPO the greedy answers still fall into loops as often as before. Of the 26 cut-off instruction-following answers, all 26 are loops. Of the 55 cut-off logic answers, 33 are loops.

\smallskip
\textbf{2. Find where the trainer sets its sampling.} The agent runs four commands to read three pieces of code: the installed trainer's code for rollouts (the files \texttt{single\_turn\_agent\_loop.py} and \texttt{agent\_loop.py} of verl), and a starter that already replaces the rollout loop for ALFWorld. It looks for two things: where the trainer passes the sampling temperature to the model, and how the trainer registers a custom rollout loop.

\smallskip
\textbf{3. Write the new rollout loop.} It subclasses the trainer's loop, with its reason in the docstring.

\begin{lstlisting}[style=agentnote]
$ cat > /workspace/rl/greedy_mix_loop.py <<'EOF'
"""Single-turn agent loop that decodes a share of the training rollouts greedily.

dev6 decodes five of its six tasks greedily, and the SFT model's greedy answers loop where its sampled ones (T 0.8) do
not: in grpo1's first 25 steps, 2-7 % of the sampled rollouts cut at RESP were repetition loops, against 55-100 % of
the greedy validation answers cut there, and 25 steps of GRPO on sampled rollouts left greedy truncation where it was.
So here each training rollout is greedy with probability GREEDY_SHARE (environment, default 0.25), the others at the
configured temperature; a greedy rollout that loops to RESP scores 0 beside its group's sampled ones and gets a
negative advantage. Validation is left as configured (Verl's val_kwargs)."""
...
GREEDY_SHARE = float(os.environ.get('GREEDY_SHARE', '0.25'))

@register('greedy_mix_agent')
class GreedyMixAgentLoop(SingleTurnAgentLoop):
    async def run(self, sampling_params, **kwargs):
        params = dict(sampling_params)
        if params.get('temperature', 0) > 0 and random.random() < GREEDY_SHARE:
            params.update(temperature=0, top_p=1.0, top_k=-1)
        return await super().run(params, **kwargs)
EOF
\end{lstlisting}

\textbf{4. Rebuild the training problems.} In the same turn the agent filters
its logic puzzles to those of at most 16 cells (6,398 remain), splits its
mathematics problems by source, and commits the workspace.

\smallskip
\textbf{5. Act through the environment.} Two \texttt{athanor} commands end the
turn: a data job that samples the new training dataset, and a stop for the
running GRPO job.

\begin{lstlisting}[style=agentnote]
$ git add -A; git commit -qm "greedy-mix agent loop, stage-2 pools"
$ athanor job start rl-mix2 --out /scratch/data/rl --note "RL stage-2 table: signal-rich sources, zebra <=16 cells" -- $ATHANOR_TRAINING_PYTHON -B prep/sample_rl.py /scratch/data/rl/rl2.parquet acereason=...:5000 sky36=...:2000 am3=...:1000 zebra=/scratch/data/rl/zebra-le16.parquet:3500 mcq=...:3000 ... --seed 2
$ athanor job stop grpo1
\end{lstlisting}

\textbf{Next wake-up.} The environment's journal line arrives in the session
(\emph{``job-grpo1 ended cancelled \ldots{} after 2h40m; its log's last step
25''}), and the agent goes on to train with the new loop.
\end{agenttrace}

\Needspace{10\baselineskip}
\subsection{Further Traces}\label{app:further-traces}

\begin{agenttrace}[breakable=false,label={tr:extrap}]{Finding (1): scaling a merge away from the base (\agent{3}, hours 55--56)}
\small
\textbf{Evidence.} The team's best model is \agent{4}'s weighted average of three
agents' models (63.28). \agent{3} measures how far each merge has moved from the
base and tries scaling that distance.

\smallskip
\textcolor{athanorblue}{\textsc{Researcher}}\quad\emph{ext11-s3Lm (base +
1.1*(s3Lm - base) \ldots) FULL \ldots: 63.97 \ldots{} vs linear s3Lm 63.28. \ldots{} ext12-s3Lm (x1.2 \ldots): 64.58 \ldots{} Zebra
rises with scale 56.4/59.6/63.0. \ldots{} ext13-s3Lm (x1.3) partial: mmlu 63.8,
math 35.8, zebra 61.6 -> past the peak \ldots{} Cancelled x1.5 eval.}

\smallskip
\textbf{Action.} \agent{3} keeps the factor 1.2 and evaluates it on three more seeds; its mean over the four is 63.97 (the same as factor 1.1's single seed), below the 64.58 of its first seed.  The full evaluation at
factor 1.3, finished later, gave 64.87. \agent{4} answers in its shared notes, \emph{``great find, r3. Running
(not duplicating yours): ext13-s3Lo''}, and reaches 65.51 with later parents. The scores at factors 1.1 to 1.3 (63.97 to 64.87) differ by about as much as one seed's noise (\cref{fig:c1-science}a).
\end{agenttrace}

\begin{agenttrace}[breakable=false,label={tr:rft}]{Finding (1): designing the self-distillation rounds (\agent{3}, hours 57--66)}
\small
\textbf{Evidence.} After the extrapolated merge, \agent{3} reads which answers
of its best model still fail.

\smallskip
\textcolor{athanorblue}{\textsc{Researcher}}\quad\emph{ext12 eval: math cut at
16k = 240/732 (33\%), none of them loops \ldots: genuine long, wandering
reasoning. Zebra cut 87/500. Biggest gaps vs official: zebra -20, math -10.
Plan: expert iteration / RFT from the best soup (allowed: samples of the model
we train, kept by programs): sample soup on zebra + math (+code/mcq) prompts at
16k \ldots, keep verified-correct, prefer shortest; SFT the soup 1 epoch low
lr; eval; iterate / soup with parent / rescale.}

\smallskip
\textbf{Action.} In the first round the agent keeps the shortest correct answer to each problem, 4,840 answers in all, and trains the model on them. The model's development score rises from 63.97 (four seeds) to 64.99 (three seeds). \agent{2} copies the
programs the same day and uses \agent{3}'s samples to fine-tune \agent{4}'s merge.
\end{agenttrace}

\begin{agenttrace}[breakable=false,label={tr:soup}]{Finding (2): averaging two agents' reinforcement-learning models (\agent{4}, hours 17--21)}
\small
\textbf{Evidence.} \agent{1} and \agent{3} each have a reinforcement-learning model near 57,
and \agent{4} averages the two.

\smallskip
\textcolor{athanorblue}{\textsc{Researcher}}\quad\emph{soup of the two
$\sim$57 RL models = 59.95 full dev6 \ldots{} every task but alfworld
above both parents. Lesson: averaging two EQUAL-strength RL models from
different lineages gives $\sim$+2.7; averaging a strong one with
weaker ones is neutral (s4a). Suggestion: keep RL lineages DIFFERENT (own
SFT/data/settings) and soup the strong ones again; anyone can also RL from this
soup.}

\smallskip
\textbf{Action.} This merge, at hour 21, becomes the team's best checkpoint. It is neither \agent{4}'s hour-15.9 merge in \cref{tab:team-events} nor \agent{1}'s average of $R_1$ and $R_3$ in \cref{tab:lineage-steps}, and it is not on the submission's lineage.
\end{agenttrace}

\begin{agenttrace}[breakable=false,label={tr:lineageB}]{Finding (3): a lineage without the shared corpus (\agent{3}, hours 36--46)}
\small
\textbf{Evidence.} Every fine-tuned model of the team is built on the same
corpus of reasoning traces. \agent{3} notices and trains a second lineage on
other data.

\smallskip
\textcolor{athanorblue}{\textsc{Researcher}}\quad\emph{All team SFTs so far
(mine, r1's, r2's v1/v2) are AM-Thinking based. For soup diversity I am
training lineage B from base on [DeepSeek-]R1 traces instead \ldots{}} Ten hours later:
\emph{Lineage B stuck $\sim$55 \ldots{} soup .67 r1-soupe + .33 B2s10:
mmlu 66.2, math 32.7, zebra 44.6 (soup-e 53.8): lineage B dilutes zebra ->
dropped B.}

\smallskip
\textbf{Action.} \agent{3} goes back to the lineage that all four agents share.
\end{agenttrace}

\begin{agenttrace}[breakable=false,label={tr:solver}]{Finding (4): rejecting a data source after reading model outputs (\agent{2}, hours 31--32)}
\small
\textbf{Evidence.} \agent{2} writes a solver producing step-by-step traces
for logic puzzles and trains on 6,900 of them.

\smallskip
\textcolor{athanorblue}{\textsc{Researcher}}\quad\emph{zebra solver-trace SFT
(\ldots 6.9k Emulated-Inc solver traces + 1.9k replay): zebra-only diag 27.0
(vs $\sim$49) -> FAILED; model mixes attributes \ldots; eval clue
phrasing differs. Dropped. \ldots{} zsft failure analysis: model imitates trace
style but asserts unjustified deductions (state not explicit in trace).}

\smallskip
\textbf{Action.} The data source is dropped after one training run and a read
of the model's answers; the agent does not train on it again.
\end{agenttrace}

\begin{agenttrace}[breakable=false,label={tr:compact}]{Finding (4): checking the data before spending GPU hours (\agent{1}, hour 95)}
\small
\textbf{Evidence.} The team's models fail the large logic puzzles. \agent{1}
considers training on short correct solutions and first checks whether any
exist among the model's own samples. The quote gives, per puzzle size, how many solved puzzles have a correct answer under 8k and under 10k tokens.

\smallskip
\textcolor{athanorblue}{\textsc{Researcher}}\quad\emph{compact-solution RFT
idea checked and dropped \ldots: shortest correct per prompt under 8k / 10k
tokens: 4x5 171/287 of 382 solved, 4x6 42/165 of 363, 5x4 106/208 of 347, 5x5
3/37 of 157, 6x4 3/31 of 121. Large puzzles need 10-16k tokens in the model's
own style; no short correct traces to learn from.}

\smallskip
\textbf{Action.} The idea is dropped without a training run.
\end{agenttrace}

\end{document}